\documentclass{article}

\usepackage[numbers]{natbib}

\usepackage[preprint]{neurips_data_2023}

\usepackage[utf8]{inputenc} 
\usepackage[T1]{fontenc}    
\usepackage{hyperref}       
\usepackage{url}            
\usepackage{booktabs}       
\usepackage{amsfonts}       
\usepackage{nicefrac}       
\usepackage{microtype}      
\usepackage{xcolor}         
\usepackage{algorithm}
\usepackage{algorithmic}
\usepackage{amsmath}
\usepackage{graphicx}
\usepackage{bm}
\usepackage{multirow}
\usepackage{wrapfig}
\usepackage{subfig}
\usepackage{url}
\definecolor{darkgreen}{rgb}{0.09, 0.45, 0.27}
\definecolor{snsorange}{RGB}{255, 141, 98}
\definecolor{codegreen}{rgb}{0,0.5,0}
\definecolor{codered}{rgb}{0.7,0.1,0.1}
\definecolor{codegray}{rgb}{0.5,0.5,0.5}
\definecolor{codepurple}{rgb}{0.58,0,0.82}
\definecolor{backcolour}{rgb}{1,1,1}
\definecolor{gred}{rgb}{0.859,0.267,0.216}
\definecolor{ggreen}{rgb}{0.059,0.616,0.345}
\definecolor{gblue}{rgb}{0.259,0.522,0.957}
\definecolor{gyellow}{rgb}{0.957,0.706,0}
\definecolor{gpurple}{rgb}{0.565,0.173,0.894}

\usepackage{ragged2e}
\usepackage{enumerate}
\usepackage{enumitem}
\usepackage{colortbl}
\usepackage{amsmath}
\usepackage{amssymb}
\usepackage{mathtools}
\usepackage{amsthm}
\usepackage{changepage}

\theoremstyle{plain}

\theoremstyle{definition}

\theoremstyle{remark}

\usepackage{bbding}
\usepackage{xspace}
\newcommand{\ourshort}{RL-ViGen\xspace}

\title{RL-ViGen: A Reinforcement Learning Benchmark for Visual Generalization}

\author{
Zhecheng Yuan$^{1,3*}$, \quad
Sizhe Yang$^{2,3}\thanks{equal contribution}$, \quad
Pu Hua$^{1,3}$, \quad
Can Chang$^{1,3}$, \quad \\
\textbf{Kaizhe Hu$^{1,3}$}, 
\vspace{0.2cm}
\textbf{
Huazhe Xu$^{1,3,4}$} \\
\vspace{0.1cm}
$^{1}$ Tsinghua University, $^{2}$ University of Electronic Science and Technology, \\  $^{3}$ Shanghai Qi Zhi Institute, $^{4}$Shanghai AI Lab\\
\texttt{yuanzc23@mails.tsinghua.edu.cn,
huazhe\_xu@mail.tsinghua.edu.cn}
\vspace{1.0cm}
}

\begin{document}

\maketitle

\begin{abstract}

  Visual Reinforcement Learning~(Visual RL), coupled with high-dimensional observations, has consistently confronted the long-standing challenge of out-of-distribution generalization. 
  Despite the focus on algorithms aimed at resolving visual generalization problems, we argue that the devil is in the existing benchmarks as they are restricted to isolated tasks and generalization categories, undermining a comprehensive evaluation of agents' visual generalization capabilities. 
  To bridge this gap, we introduce \ourshort: a novel \textbf{R}einforcement \textbf{L}earning Benchmark for \textbf{Vi}sual \textbf{Gen}eralization, which contains diverse tasks and a wide spectrum of generalization types, thereby facilitating the derivation of more reliable conclusions. 
  Furthermore, \ourshort incorporates the latest generalization visual RL algorithms into a unified framework, under which the experiment results indicate that no single existing algorithm has prevailed universally across tasks. 
  Our aspiration is that \ourshort will serve as a catalyst in this area, and lay a foundation for the future creation of universal visual generalization RL agents suitable for real-world scenarios. 
  Access to our code and implemented algorithms is provided at \url{https://gemcollector.github.io/RL-ViGen/}.
  
\end{abstract}

\section{Introduction}
Visual Reinforcement Learning~(RL) has attained remarkable success across a plethora of domains~\cite{mnih2015human, ouyang2022training, feng2023dense}. A diverse range of techniques has been implemented to tackle not only the trial-and-error learning process but also the complexity arising from high-dimensional input data. Notwithstanding these successes, a fundamental challenge confronting visual RL agents persists --- achieving generalization.

To overcome this obstacle, several visual RL generalization benchmarks have emerged, including Procgen~\cite{cobbe2020leveraging}, Distracting Control Suite~\cite{stone2021distracting}, and DMC-GB~\cite{hansen2021generalization}. While these benchmarks have been indispensable to visual RL generalization progress, they are not exempt from inherent limitations that pose challenges to further development. 
Procgen offers a diverse distribution of environment configurations and visual appearances. 
However, it is limited to video games with non-realistic images and low-dimensional discrete action spaces, resulting in a significant gap between its environments and real-world scenarios. 
Another instance, DMC-GB, is sometimes treated as a golden standard for many state-of-the-art visual generalization algorithms. 

Nevertheless, the narrow scope of task classes and generalization categories in existing setups cannot thoroughly and comprehensively evaluate the agent's generalization ability.
In addition, although Distracting Control Suite contains two generalization types, it falls short in diversity and complexity.   
The prevailing trend in this field is to showcase the superiority of proposed algorithms on these benchmarks, which adversely poses a certain risk of promoting overfitting to these benchmarks, rather than discovering algorithms potentially beneficial for solving real-world problems.

In this paper, we introduce a novel \textbf{R}einforcement \textbf{L}earning benchmark for \textbf{Vi}sual \textbf{Gen}eralization~(\ourshort), presenting numerous merits over existing counterparts. 
Our benchmark integrates a spectrum of task categories with realistic image inputs, including table-top manipulation, locomotion, autonomous driving, indoor navigation, and dexterous hand manipulation, allowing for a more comprehensive evaluation of the agents' efficacy. Moreover, by incorporating various key aspects in visual RL generalization, such as visual appearances, lighting changes, camera views, scene structures, and cross embodiments, \ourshort enables a comprehensive examination of agents' generalization ability against distinct visual conditions.

It is noteworthy that we provide a unified framework that encompasses various state-of-the-art visual RL and generalization algorithms with the same optimization scheme for each approach. The framework not only promotes fair benchmarking comparisons but also lowers the entry barrier for devising novel approaches.

\begin{figure}[t]
  \centering
  \includegraphics[width=1.0\linewidth]{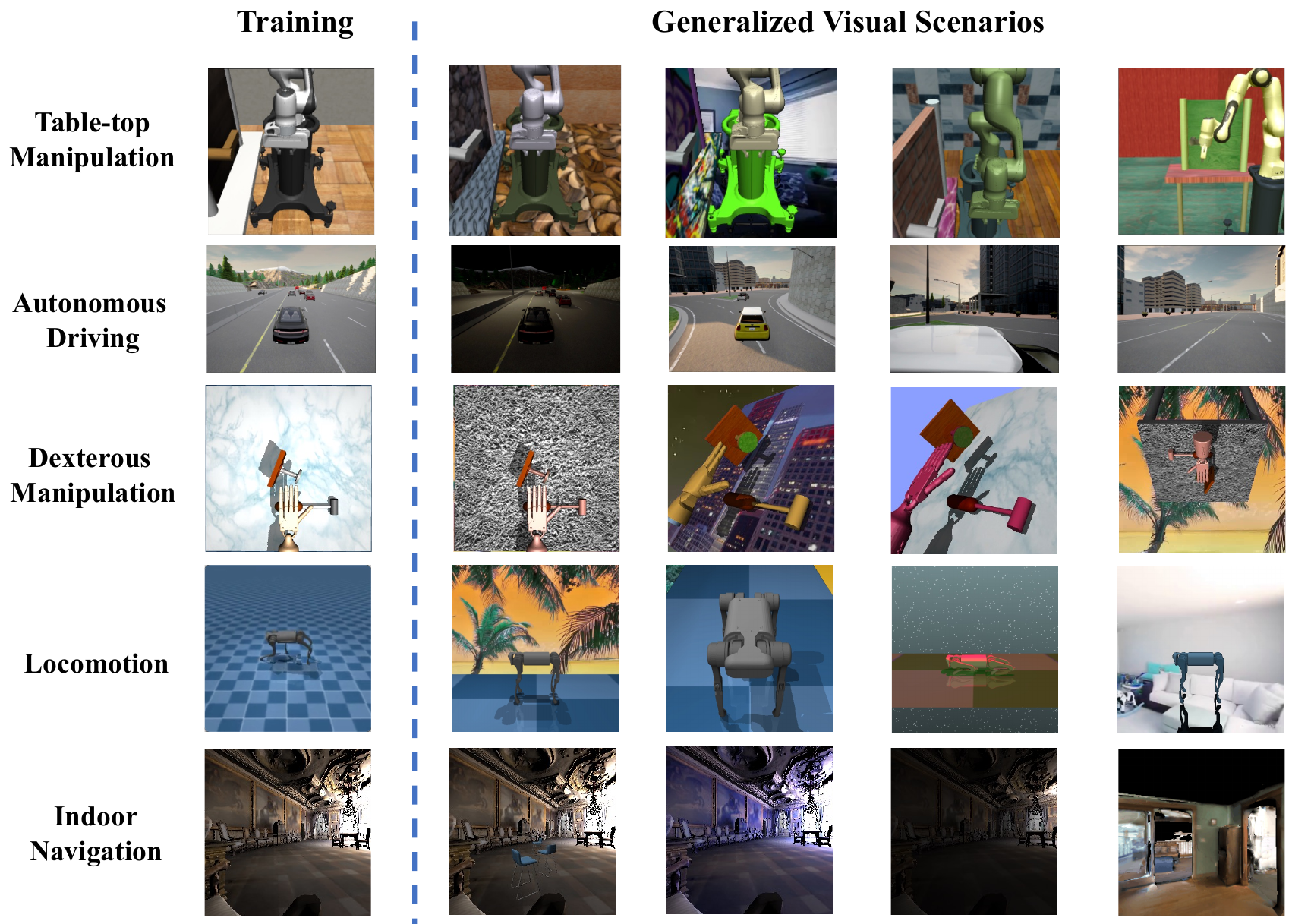}
  \caption{\textbf{The novel RL benchmark for visual generalization.} We show that \ourshort supports a wide range of tasks with different generalization categories. The algorithms can be evaluated more comprehensively and achieve more convincing experimental results.}
  \label{fig:bench}
  \vspace{-10pt}
\end{figure}

In summary, our contributions are as follows: \textbf{1)}~we propose a novel visual RL generalization benchmark~\ourshort with diverse, realistic rendering tasks and numerous generalization types;  \textbf{2)}~we implement and evaluate various algorithms within a unified framework, enabling a comprehensive analysis of their generalization performance; \textbf{3)}~we conduct comprehensive and extensive experiments to demonstrate the distinct performance of existing approaches when tackling diverse tasks and generalization types, and highlight the benefits and the limitations of current generalizable visual RL algorithms. With all the contributions combined, \ourshort may pave the way for further advancements in visual RL generalization, ultimately leading to more robust and adaptable algorithms for real-world applications.

\section{\ourshort}

\ourshort consists of 5 distinct task categories, spanning the domain of locomotion, table-top manipulation, autonomous driving, indoor navigation, and dexterous hand manipulation. In contrast to prior benchmarks, \ourshort employs a diverse array of task classes for evaluating the agent's generalization performance. We believe that only through comprehensive examination from multiple perspectives can we obtain convincing results. Furthermore, as shown in Table~\ref{table:general}, our benchmark offers a wide range of generalization categories, including visual appearances, camera views, variations in lighting conditions, scene structures, and cross embodiments settings, thereby providing a thorough evaluation of algorithms' robustness and generalization abilities.

\begin{table}[t]
    \centering
    \vspace*{-9pt}
    \captionof{table}{\textbf{Generalization Categories.} The following table outlines the types of generalization incorporated within each task. Except for categories considered as not applicable~(N/A) (e.g., for locomotion, changes in scene structures are not required), all potential types are included.}
    \vspace{0.075in}
    \label{table:general}
    \renewcommand\tabcolsep{4.0pt}
    \begin{tabular}[b]{cccccc}
    \toprule
    \begin{tabular}[c]{@{}c@{}}\textbf{Generalization},\\ \textbf{Categories}\end{tabular} & \begin{tabular}[c]{@{}c@{}}Visual\\ Appearances\end{tabular} & \begin{tabular}[c]{@{}c@{}}Camera\\ Views\end{tabular} & \begin{tabular}[c]{@{}c@{}}Lighting \\ Changes\end{tabular}  & \begin{tabular}[c]{@{}c@{}}Scene \\ Structures \end{tabular} & \begin{tabular}[c]{@{}c@{}}Cross \\ Embodiments\end{tabular}    \\
    \midrule
    \begin{tabular}[c]{@{}c@{}} \underline{{\textbf{Autonomous Driving}}} \end{tabular} & \begin{tabular}[c]{@{}c@{}} \textcolor{ggreen}{\large{\Checkmark}} \end{tabular} & \begin{tabular}[c]{@{}c@{}} \textcolor{ggreen}{\large{\Checkmark}} \end{tabular} & \begin{tabular}[c]{@{}c@{}} \textcolor{ggreen}{\large{\Checkmark}} \end{tabular} & \begin{tabular}[c]{@{}c@{}}\textcolor{ggreen}{\large{\Checkmark}} \end{tabular} & \begin{tabular}[c]{@{}c@{}}\textcolor{ggreen}{\large{\Checkmark}} \end{tabular} \\
    \midrule
    \begin{tabular}[c]{@{}c@{}}\underline{{\textbf{Table-top Manipulation}}} \end{tabular} & \begin{tabular}[c]{@{}c@{}} \textcolor{ggreen}{\large{\Checkmark}} \end{tabular} & \begin{tabular}[c]{@{}c@{}} \textcolor{ggreen}{\large{\Checkmark}} \end{tabular} & \begin{tabular}[c]{@{}c@{}} \textcolor{ggreen}{\large{\Checkmark}} \end{tabular} & \begin{tabular}[c]{@{}c@{}} \textbf{N/A}                           \end{tabular} & \begin{tabular}[c]{@{}c@{}} \textcolor{ggreen}{\large{\Checkmark}} \end{tabular} \\
    \midrule
    \begin{tabular}[c]{@{}c@{}} \underline{{\textbf{Indoor Navigation}}}   \end{tabular} & \begin{tabular}[c]{@{}c@{}} \textcolor{ggreen}{\large{\Checkmark}} \end{tabular} & \begin{tabular}[c]{@{}c@{}} \textcolor{ggreen}{\large{\Checkmark}} \end{tabular} & \begin{tabular}[c]{@{}c@{}}  \textcolor{ggreen}{\large{\Checkmark}}                                     \end{tabular} & \begin{tabular}[c]{@{}c@{}} \textcolor{ggreen}{\large{\Checkmark}} \end{tabular} & \begin{tabular}[c]{@{}c@{}}      \textbf{N/A}                                    \end{tabular} \\
    \midrule
    \begin{tabular}[c]{@{}c@{}} \underline{{\textbf{Dexterous Manipulation}}} \end{tabular} & \begin{tabular}[c]{@{}c@{}} \textcolor{ggreen}{\large{\Checkmark}} \end{tabular} & \begin{tabular}[c]{@{}c@{}} \textcolor{ggreen}{\large{\Checkmark}} \end{tabular} & \begin{tabular}[c]{@{}c@{}} \textcolor{ggreen}{\large{\Checkmark}} \end{tabular} & \begin{tabular}[c]{@{}c@{}}  \textbf{N/A}   \end{tabular} & \begin{tabular}[c]{@{}c@{}}  \textcolor{ggreen}{\large{\Checkmark}} \end{tabular} \\
    \midrule
    \begin{tabular}[c]{@{}c@{}} \underline{{\textbf{Locomotion}}} \end{tabular} & \begin{tabular}[c]{@{}c@{}} \textcolor{ggreen}{\large{\Checkmark}} \end{tabular} & \begin{tabular}[c]{@{}c@{}} \textcolor{ggreen}{\large{\Checkmark}} \end{tabular} & \begin{tabular}[c]{@{}c@{}} \textcolor{ggreen}{\large{\Checkmark}} \end{tabular} & \begin{tabular}[c]{@{}c@{}}         \textbf{N/A}                              \end{tabular} & \begin{tabular}[c]{@{}c@{}} \textcolor{ggreen}{\large{\Checkmark}} \end{tabular} \\
    \bottomrule
    \end{tabular}
    \vspace{-0.2in}
\end{table}

\subsection{Environments}
\textbf{Dexterous manipulation:}
Adroit~\cite{rajeswaran2017learning} is a sophisticated environment that is explicitly tailored for dexterous hand manipulation tasks. 
It demands considerable exploration and fine-grained feature capturing due to the sparse reward nature of the environment and the complexity of high-dimensional action space. 
In \ourshort, we have enriched the Adroit environment by integrating diverse visual appearances, camera perspectives, hand types, lighting changes, and object shapes.

\textbf{Autonomous driving:}
CARLA~\cite{dosovitskiy2017carla} serves as a realistic and high-fidelity simulator for autonomous driving, which investigates the control capabilities of agents under dynamic conditions. It has been successfully deployed on visual RL settings in prior studies. 
Contrary to previous work~\cite{CarlaEnv}, \ourshort  provides an enhanced range of dynamic weather and more complex road conditions in different scene structures. 
Furthermore, flexible camera angle adjustments are also included within  \ourshort.

\textbf{Indoor navigation:}
As an efficient and photorealistic 3D simulator, Habitat~\cite{habitat19iccv} combines numerous visual navigation tasks. Succeeding in these tasks requires the agents to own the capability of scene understanding. \ourshort builds upon the  \textit{skokloster-castle} scene and proposes additional scenarios with different visual and lighting settings. In addition, the camera view and scene structure are designed to be adjustable. 

\textbf{Table-top manipulation:}
Robosuite~\cite{zhu2020robosuite} is a modular simulation platform designed to support robot learning. It inherently contains interfaces designed to adjust various scene parameters. Recent work~\cite{pmlr-v139-fan21c} has leveraged this platform to test the agent's generalization ability of visual background changes. \ourshort further incorporates dynamic backgrounds, adaptive lighting conditions, and options for embodiment variation, refining the simulation to be closer to the real world.

\textbf{Locomotion:}
DeepMind Control is a popular continuous visual RL benchmark. DMC-GB~\cite{hansen2021softda} is developed on it and has become a widely used benchmark for evaluating generalization algorithms. Building upon DMC-GB, \ourshort introduces objects and corresponding tasks from sophisticated real-world locomotion and manipulation applications, such as the Unitree, Anymal quadrupedal robots, and the Franka Arm. What's more, \ourshort also offers a variety of generalization categories to further enrich this environment. 

More detailed implementations and modifications can be found in Appendix~\ref{append:details} and our codebase.

\begin{figure}[t]
  \centering
  \includegraphics[width=0.9\linewidth]{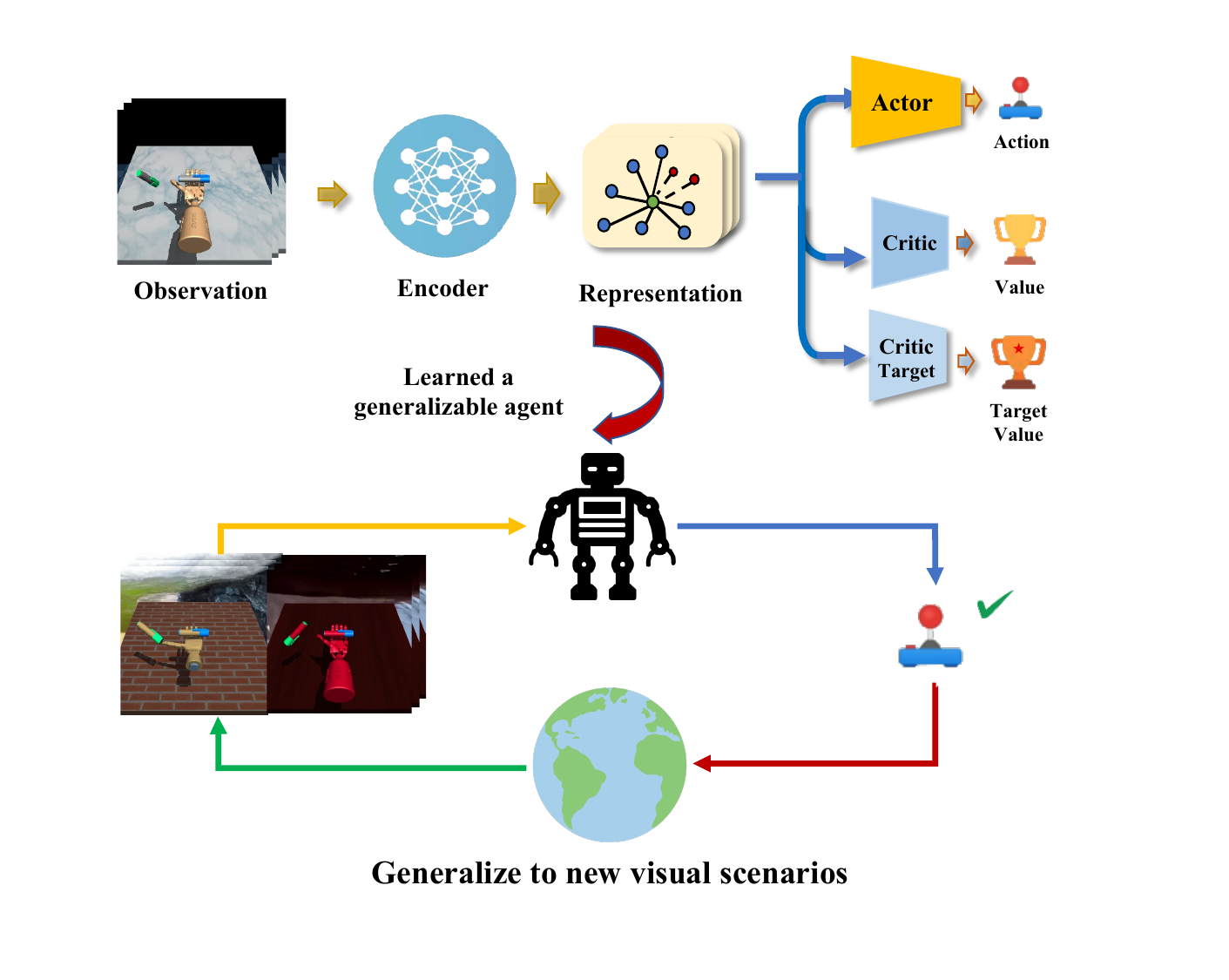}
  \caption{\textbf{Generalization procedure.} The agent is first trained in Stage 1 with a certain fixed scenario. Subsequently, in Stage 2, the agent is tested across various visual generalization scenes in a zero-shot manner. The better the agent performs in various scenes of Stage 2, the stronger generalization ability it demonstrates.}
  \label{fig:method}
  \vspace{-15pt}
\end{figure}

\subsection{Generalization Categories}

Here, we emphasize the primary generalization categories utilized in \ourshort:

\textbf{Visual appearances:} Maintaining effective performance in the presence of altered visual features of objects, scenes, or environments is of vital importance, particularly for visual reinforcement learning. In our benchmark, different components within the environment can be modified with a wide range of colors. Meanwhile, the dynamic video background is also introduced as a challenging setting.

\textbf{Camera views:} In the real world, the agents have to cope with camera configurations, angles, or positions that may not align with those experienced during training. We offer access to set the cameras at different angles, distances, and FOVs. In addition, the number of cameras can be adjusted accordingly. 

\textbf{Lighting conditions:} The change in lighting conditions will occur inevitably in the real world. To equip agents with the ability to adapt to such variations, our benchmark supplies interfaces related to the lighting, such as varied light intensity, colors, and dynamical shadow changes. 

\textbf{Scene structures:} Mastering the ability of understanding and adapting to different spatial arrangements and organization patterns within various scenes is crucial for a truly generalizable agent. To this end, our benchmark enables modifications in scene structure via adjusting maps, patterns, or introducing extra objects. 

\textbf{Cross embodiments:} Adapting learned skills and knowledge to different physical morphologies or embodiments is essential for an agent to perform well across various platforms or robots with different kinematic structures and sensor configurations. Therefore, our benchmark also provides access to modify the embodiment of trained agents in the aspects of model types, sizes, and other physical properties.

\section{Algorithmic Baselines for Generalization in Visual RL}

\subsection{A Unified Framework}
Another key contribution of our work is the implementation of a unified codebase to support comparison among various visual RL algorithms. In previous studies, different algorithms adopt distinct optimization schemes, RL baselines, and  hyperparameters. For example, SRM~\cite{huang2022spectrum} and SVEA~\cite{hansen2021stabilizing} rely on SAC-based RL algorithms, while PIE-G~\cite{yuan2022pre} utilizes a DDPG-based approach. Moreover, minor different implementations could substantially impact the final performance. Therefore, providing a unified framework is of great importance in this domain, enabling more persuasive conclusions to be drawn from evaluating algorithms across a consistent framework and diverse tasks.

\subsection{Visual RL Algorithms}
In our benchmark, we assemble eight leading visual RL algorithms and apply the same unified training and evaluation framework. \textbf{DrQ-v2}~\cite{yarats2021mastering} is the prior state-of-the-art DDPG-based model-free visual RL algorithm in terms of sample efficiency. \textbf{DrQ}~\cite{kostrikov2020image} is another SAC-based sample efficient visual RL algorithm, which is the base of DrQ-v2. \textbf{CURL}~\cite{laskin2020curl} utilizes a SimCLR-style~\cite{chen2020simple} contrastive loss to obtain better visual representations. \textbf{VRL3}~\cite{wang2022vrl3} is the state-of-the-art algorithm in Adroit tasks with human demonstrations. The other four algorithms concentrate on achieving robust representations. \textbf{SVEA}~\cite{hansen2021stabilizing} employs the Q-value of un-augmented images as the target objective while utilizing data augmentation for reducing the Q-variance; \textbf{SRM}~\cite{huang2022spectrum} adopts augmentation in the frequency domain to selectively eliminate a part of the observation frequency; \textbf{PIE-G}~\cite{yuan2022pre} incorporates ImageNet~\cite{deng2009imagenet} pre-trained model to further boost the generalization ability; \textbf{SGQN}~\cite{bertoin2022look} identifies critical pixels for decision-making via integrating with the saliency map.

\section{Experiments}
In this section, we try to investigate the generalization ability of different approaches in the proposed \ourshort benchmark. As shown in Figure~\ref{fig:method}, all agents are trained in the same fixed training environment and evaluated within various unseen scenarios in a zero-shot manner. The training sample efficiency and asymptotic performance are shown in Appendix~\ref{appendix: sample}. For each task, we evaluate over 5 random seeds and report the mean scores and 95\% confidence intervals. In terms of each trained environment, we present the aggregated scores of the multiple subtasks. The detailed and extensive experimental results can be found in Appendix~\ref{append:details} and~\ref{append:add_results}. The visualization of each environment and generalization types are shown in Appendix~\ref{append:visualization}.

\subsection{Visual Appearances and Lighting Changes}

\subsubsection{Indoor Navigation}

\begin{wrapfigure}[15]{r}{0.50\textwidth}%
    \centering
    
    \includegraphics[width=0.40\textwidth]{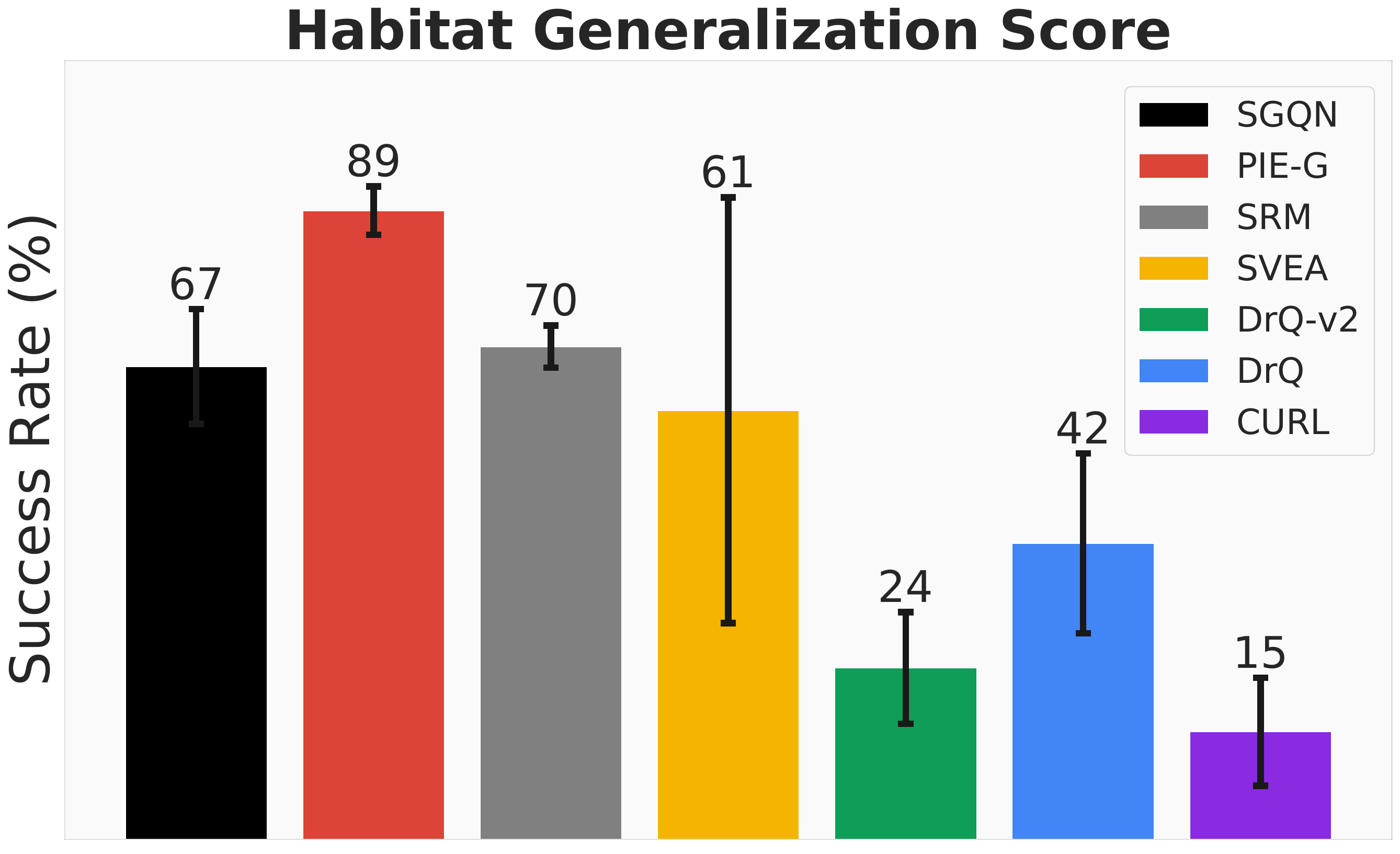}
    \caption{\textbf{Generalization score of indoor navigation.} We present the success rate of each method. The result indicates that PIE-G achieves better generalization performance on Habitat.}
    \label{fig:habi_score}
\end{wrapfigure}

Within the Habitat platform, we choose the \emph{ImageNav} task and modify the 3D scanned models to introduce novel scenarios with various visual appearances and lighting conditions. We conduct 10 evaluations in each of the 10 selected scenarios~(100 trials in total). In contrast to most existing benchmarks, the Habitat-rendered images are captured from a first-person viewpoint by the high-performance 3D simulator. Hence, it can deliver a visualization more akin to real-world scenes. As shown in Figure~\ref{fig:habi_score}, the superior performance of PIE-G can be attributed to the integration with the ImageNet pre-trained model, equipping PIE-G with a wealth of authentic images and enabling it to handle these scenarios more efficiently. Conversely, consistent with the conclusion drawn from Section~\ref{subsub:drive}, SGQN, which intends to segment the centered agent via eliminating the redundant background, is proved ineffective in these object-rich and first-person view tasks.

\subsubsection{Autonomous Driving}\label{subsub:drive}
Regarding CARLA, we adopt the reward function setting in Zhang et al.~\cite{zhang2020learning} and apply a first-person perspective to better resemble real-world driving conditions. As shown in Figure~\ref{fig:carla_app} in Appendix~\ref{append:visualization}, this environment is divided into three levels: \textit{Easy}, \textit{Medium}, and \textit{Hard}. The main modifications involve varying factors such as rainfall intensity, road wetness, and lighting. The higher the disparity from the training scenarios, the more challenging the difficulty level. In this task, one of its distinctive features is that the input image frequency undergoes considerable changes. Consequently, the SRM approach, which applies data augmentation in the frequency domain, demonstrates the best performance as it can adapt to the input images with varying frequencies. While PIE-G incorporates the ImageNet pre-trained model, its source training images mainly possess higher-frequency features, thus suffering from suboptimal generalization when facing low-frequency scenarios~(e.g., dark night). Moreover, SGQN, which extracts salient information, exhibits a decrease in performance when faced with visually rich scenes where the controlled agent is not present in the observed frame. It also should be noted that DrQ gains a degree of generalization ability in this environment. Our observations suggest that since DrQ is a SAC-based algorithm, it tends to be prone to entropy collapse~\cite{yarats2021mastering}. This implies that the trained agent only produces a single distribution of action in response to diverse image inputs.

\begin{figure}[h]
  \centering
  \includegraphics[width=0.9\linewidth]{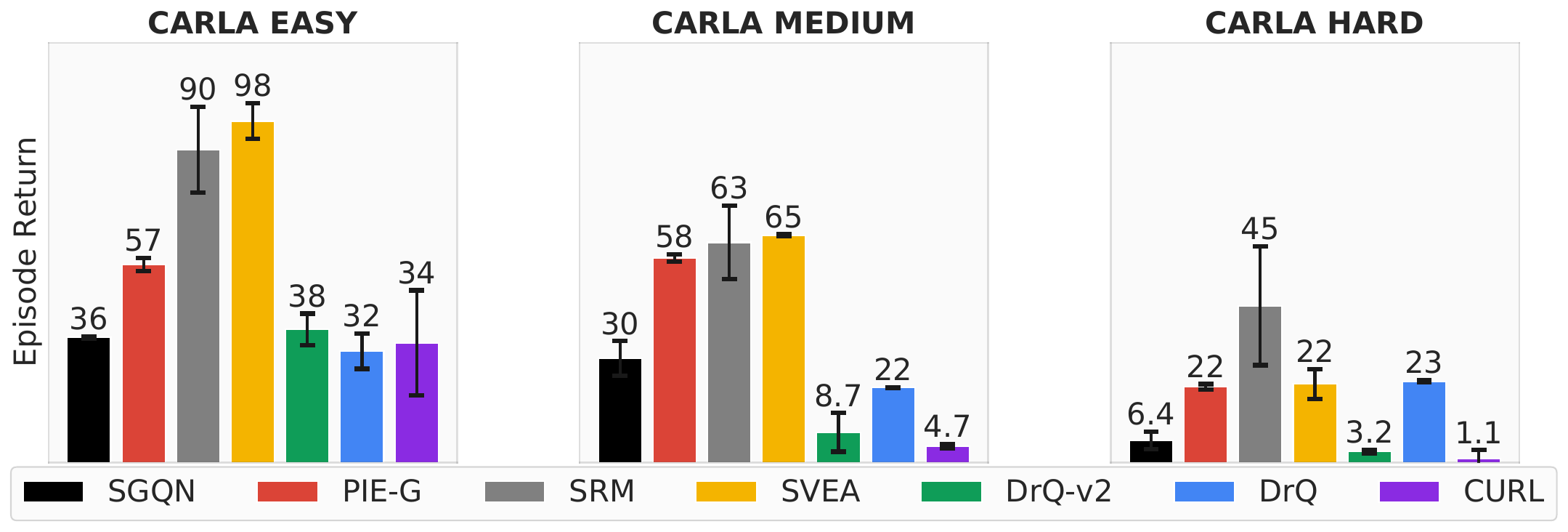}
  \caption{\textbf{Aggregated generalization score of autonomous driving.} We present the aggregated return of each method. SRM exhibits better performance to adapt to scenarios where image frequency varies dramatically.}
  \label{fig:carla}
\end{figure}

\subsubsection{Dexterous Hand Manipulation}\label{subsub:dexterous}
In the Adroit environment, we assess the performance of each approach in three single-view tasks: \textit{Door}, \textit{Hammer}, and \textit{Pen}. Since DrQ-v2 and DrQ barely perform well in these challenging environments, we utilize VRL3~\cite{wang2022vrl3}, the state-of-the-art method in this domain, as the base algorithm and the visual RL approaches in \ourshort are re-implemented upon it.

 With respect to sample efficiency, it is commonly believed that applying strong augmentation will negatively affect sample efficiency. However, as illustrated in Figure~\ref{fig:adroit_curve} in Appendix~\ref{appendix: sample}, it is worth noting that since VRL3 specifically designs a safe Q mechanism to prevent potential Q divergence for this environment, the generalization algorithms applying strong augmentation can achieve performance comparable to those using only random shift. 

 \begin{figure}[h]
  \centering
  \includegraphics[width=1.0\linewidth]{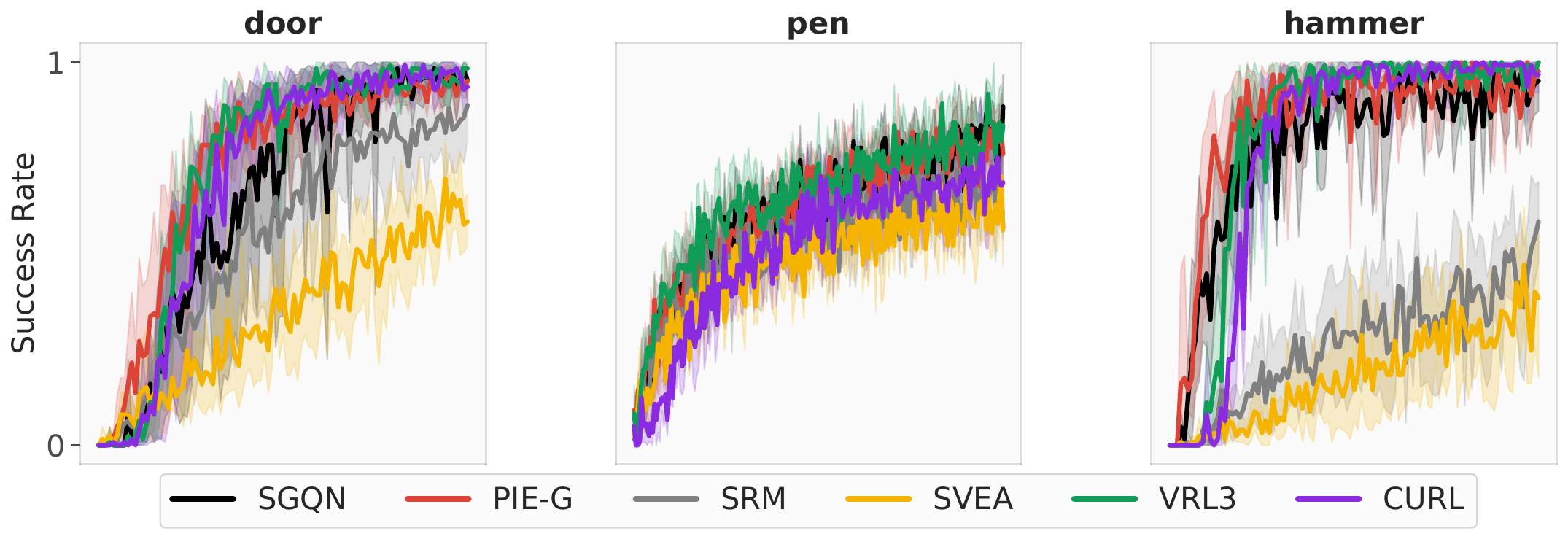}
  \caption{\textbf{Sample efficiency of Adroit.} The success rate of each algorithm. We normalize the training steps into (0, 1). The approaches with strong augmentation can also gain comparable performance. }
  \label{fig:adroit_curve}
\end{figure}

\begin{figure}[t]
  \centering
  \includegraphics[width=1.0\linewidth]{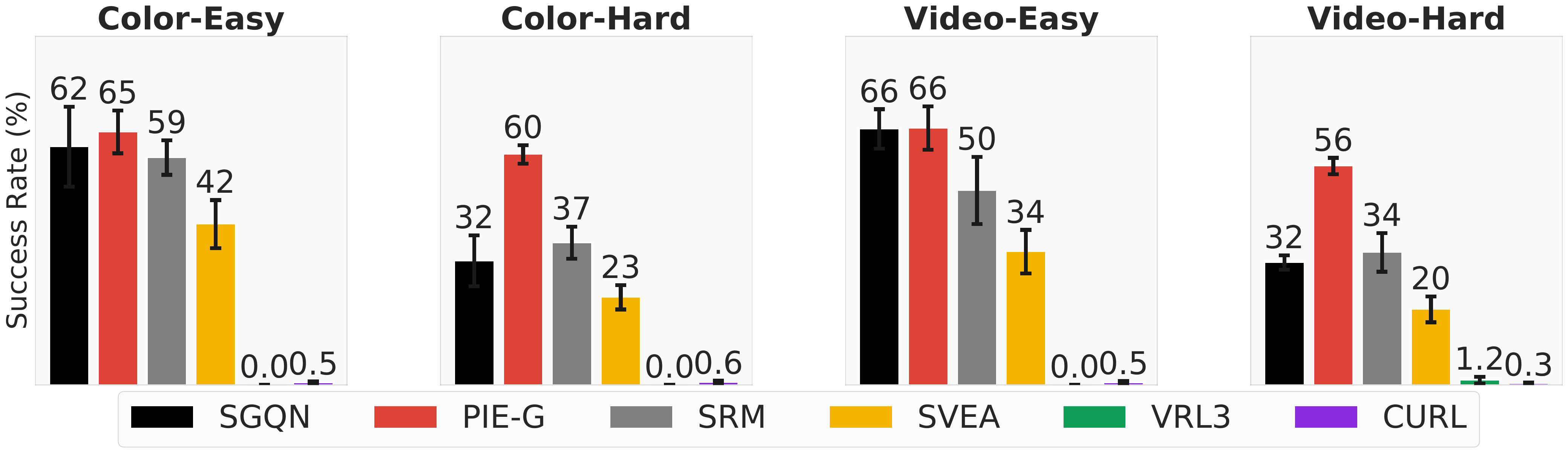}
  \caption{\textbf{The aggregated generalization score of dexterous manipulation.} We present the aggregated success rate of each method. PIE-G equipped with the ImageNet pre-trained model exhibits better adaptability to Adroit tasks which necessitate fine-grained information capture.}
  \label{fig:adroit_score}
    \vspace{-15pt}
\end{figure}

As for generalization, Adroit tasks require agents to identify fine-grained features for dexterous and sophisticated manipulation. Therefore, PIE-G, which leverages ImageNet pre-trained models to capture detailed information, demonstrates the effectiveness of assisting the learned agent in executing downstream tasks, particularly in the \textit{hard} setting. Moreover, as illustrated in Figure~\ref{fig:adroit_score}, the absence of additional objectives to mitigate the effect of visual changes causes both VRL3 and CURL to struggle in adapting to novel visual situations in these demanding tasks.

\subsection{Scene Structures}
\begin{wrapfigure}[15]{r}{0.45\textwidth}%
    \centering
    \vspace{-15pt}
    \includegraphics[width=0.40\textwidth]{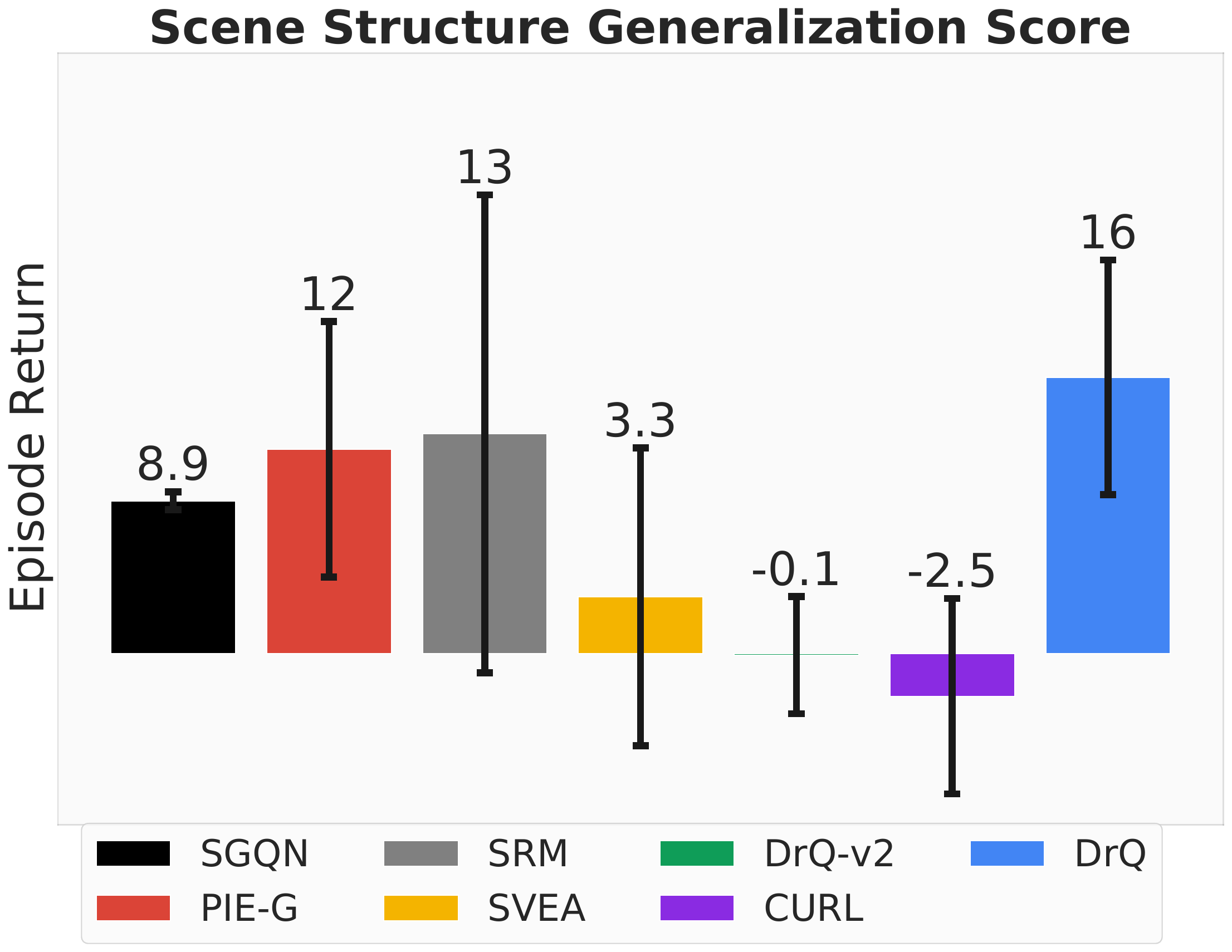}
    \vspace{-6pt}
    \caption{\textbf{Generalization score of Scene Structure.} Across this category of generalization, all algorithms demonstrate unsatisfactory performance.}
    \label{fig:scene_score}
\end{wrapfigure}

Generalizable agents that are capable of delivering robust performance across diverse scene structures are essential for potential broad real-world applications. We select CARLA as the testbed for evaluating the generalization of scene structures. The agents are trained in standard training scenarios~(highways), and tested in more complex structure settings, including narrow roads, tunnels, and roundabouts with \emph{HardRainSunset} weather conditions. As shown in Figure~\ref{fig:scene_score}, the performance of all algorithms falls short of expectations,  suggesting that the current visual RL algorithms and generalization approaches are not adequately robust to scene structural changes. More in-depth investigations must be pursued in order to enhance the generalization ability of trained agents to perceive the changing scene structures.

\subsection{Camera Views}
\begin{wrapfigure}[13]{r}{0.75\textwidth}%
    \centering
    \vspace{-17pt}
    \includegraphics[width=0.70\textwidth]{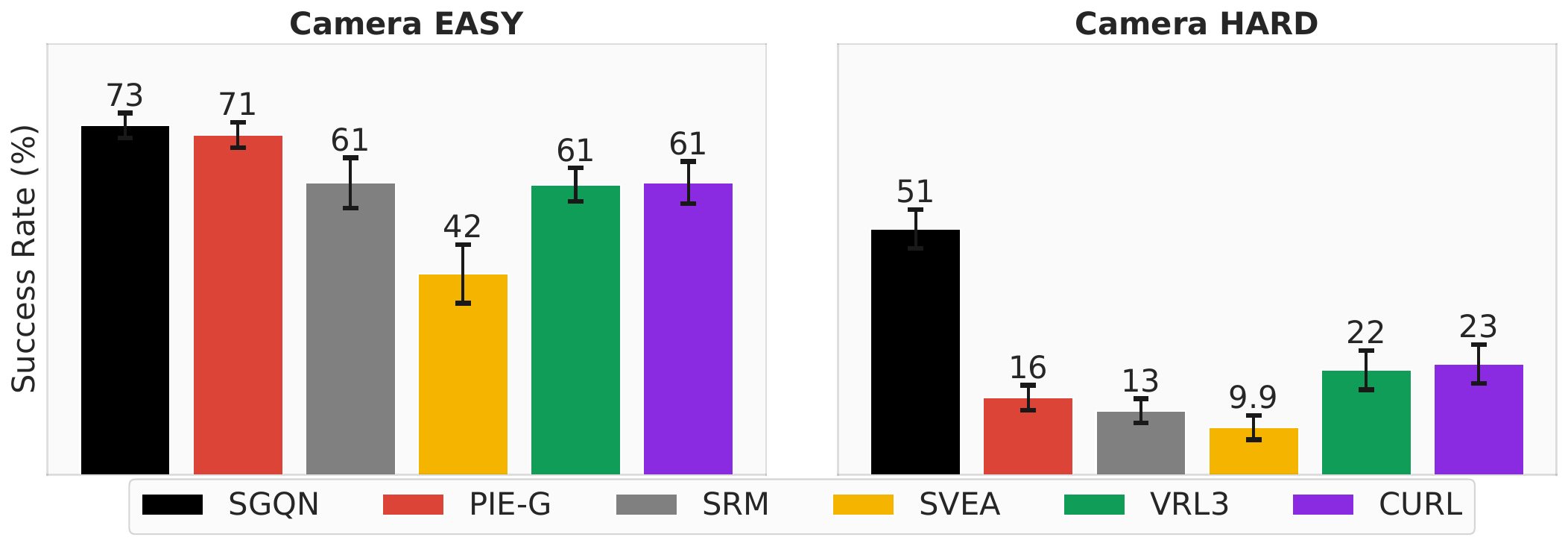}
    \caption{\textbf{Generalization score of Camera Views.} SGQN indicates advantageous generalization ability across different levels in camera-view generalization.}
    \label{fig:cam_score}
\end{wrapfigure}

We proceed to evaluate the generalization in terms of camera views in the Adroit Environment. As illustrated in Figure~\ref{fig:cam_score}, under the \textit{Easy} setting, PIE-G and SGQN exhibit leading generalization capabilities with respect to camera view, while other algorithms also demonstrate some degree of generalization due to the use of random shift augmentation. However, in the \textit{Hard} setting, which introduces substantial changes in camera position, orientation, and field of view (FOV), nearly all algorithms, except for SGQN, lose their generalization ability. The exceptional performance of SGQN is mainly due to its heavy reliance on producing saliency maps, which enhances the agent's self-awareness of object geometry and relative positioning. Hence, this property strengthens its generalization performance even in the face of major camera view alterations.

\subsection{Cross Embodiments}
\begin{wrapfigure}[16]{r}{0.55\textwidth}%
    \centering
    \vspace{-15pt}
    \includegraphics[width=0.40\textwidth]{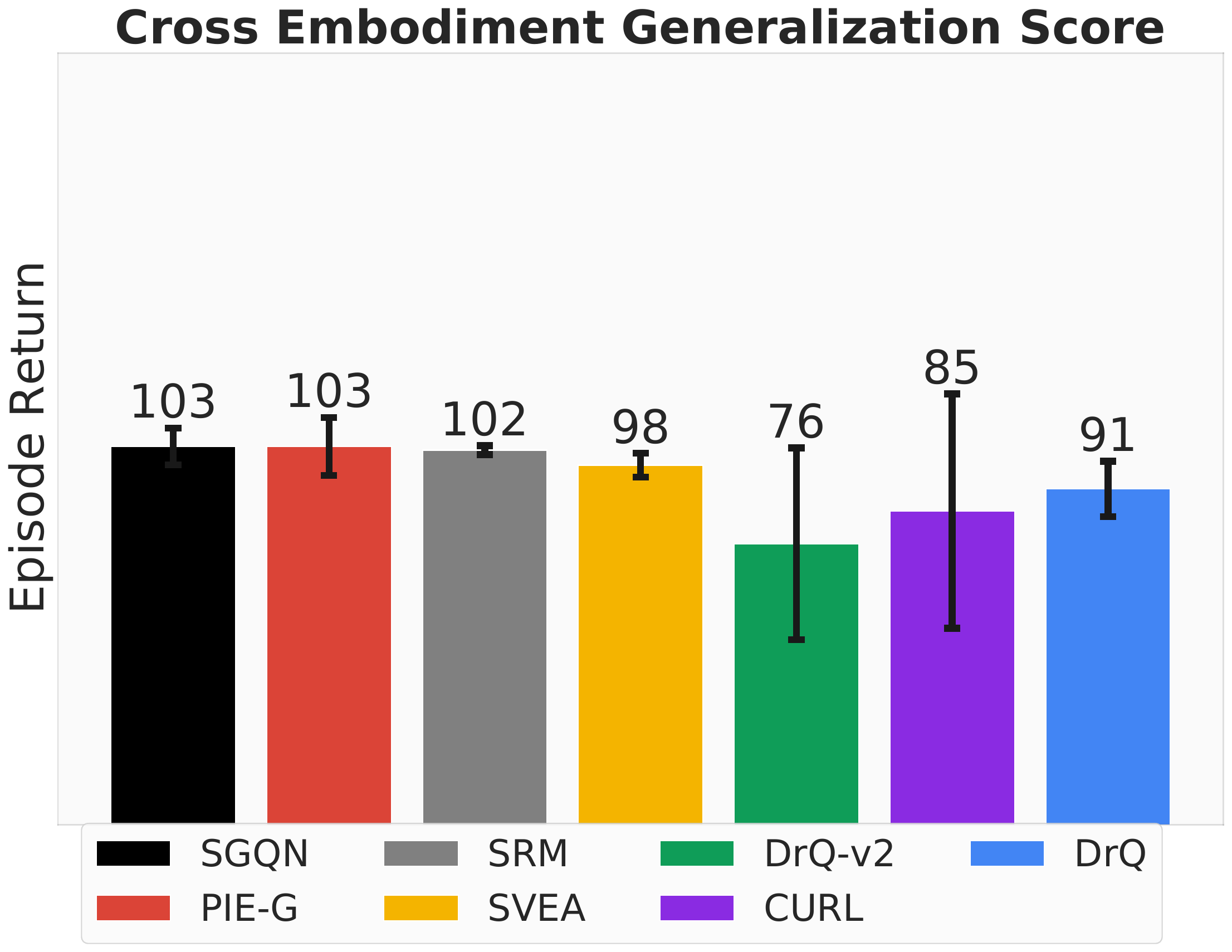}
    \vspace{-3pt}
    \caption{\textbf{The aggregated generalization score of Cross Embodiments.} No algorithm has demonstrated the capability to manage the cross-embodiment generalization yet.}
    \label{fig:cross_score}
\end{wrapfigure}

Addressing the embodiment mismatch from visual input is crucial, as the embodiment composes a substantial portion of the image and significantly influences robot behavior of interacting with the world. To investigate this type of generalization, Robosuite is employed as the evaluation platform. We utilize the OSC\_POSE controller~\cite{nakanishi2008operational} during training to facilitate the maintenance of action space dimensions and their respective meanings. Then, the trained agents transfer from Panda Arm to two different morphologies: KUKA IIWA and Kinova3. As illustrated in Figure~\ref{fig:cross_score}, the overall performance of all algorithms is suboptimal; however, generalization-based methods, which contain more diverse information during training, exhibit a slight advantage over those primarily focused on sample efficiency in the cross-embodiment setting. 

\section{Discussion}

In summary, our experiments reveal that the findings based on previous benchmarks may not accurately reflect the actual progress, leading to a distorted perception of the situation; those advanced visual RL algorithms, previously perceived as cutting-edge, display less efficacy within \ourshort. We summarize the main takeaways as follows: 

\textbf{Takeaway 1.} The experimental results reveal the varying generalization performance of different visual RL algorithms in distinct tasks and generalization categories, with no single algorithm demonstrating universally strong generalization abilities.

\textbf{Takeaway 2.} Solely enhancing training performance fails to guarantee an improvement in the generalization ability. Although DrQ(v2) and CURL exhibit high sample efficiency during training and even achieve better asymptotic performance~(Appendix~\ref{appendix: sample}), their performance in various generalization scenes has yet to reach a satisfactory level. Therefore, when attempting to improve the generalization ability of an agent, it is crucial to introduce additional inductive biases to aid the training process.

\textbf{Takeaway 3.} An effective generalizable visual RL agent must demonstrate exceptional performance across multiple generalization categories. Previous work has primarily focused on generalization concerning visual appearances, while our experiments reveal considerable shortcomings of existing algorithms in the setting of cross embodiments and scene structures. These underperforming generalization categories go beyond altering the observation space within the Markov Decision Process~(MDP); they also bring modifications to the action space and transition probabilities, thus presenting the agent with extra challenges.

\textbf{Takeaway 4.} Each generalization algorithm possesses its own unique strengths. Notably, PIE-G demonstrates superior performance with respect to visual appearances and lighting condition changes, while SRM, under significant image frequency variations, exhibits remarkable robustness. SGQN retains its generalization capacity when facing considerable camera view alterations. In addition, SVEA, without the need for additional parameters and with only minimal modifications, can achieve a certain level of generalization abilities. We hypothesize that stronger performance might be attained through a fusion of different algorithms, such as utilizing pre-trained models with frequency-based augmentation to induce further improvement.

Combined with the takeaways, we hope that an algorithm's success in \ourshort can indicate its potential applicability in more complex and unpredictable real-world scenarios. In the future, a holistic and multi-dimensional approach, encompassing aspects such as scene structures, camera views, and cross embodiments, is critical for fostering truly generalizable agents capable of navigating in varied and dynamic real-world environments. Equally, the design of more sophisticated and realistic training environments which enable to reflect the complexity of real-world conditions, can also serve as a crucial area for future explorations.

\section{Related Work}
\paragraph{RL benchmarks.} There exists a multitude of mature benchmarks aiming for evaluating reinforcement learning algorithms~\cite{chendaxbench,wang2019benchmarking,majumdar2023we,duan2016benchmarking,gu2023maniskill2,mu2021maniskill,Xiang_2020_SAPIEN}. For instance, Atari~\cite{bellemare2013arcade} and Gym-MuJoCo~\cite{todorov2012mujoco} are exemplary benchmarks in deep reinforcement learning. In other subdomains, D4RL~\cite{fu2020d4rl} serves as a popular benchmark for offline RL algorithms, while URLB~\cite{laskin2021urlb} provides an evaluating platform with respect to unsupervised RL algorithms. MetaWorld~\cite{yu2020meta} is often used to evaluate multi-task and meta-learning scenarios. SafetyGym~\cite{Ray2019}, meanwhile, is predominantly applied for testing Safe RL algorithms. Recently, 
MineDojo~\cite{fan2022minedojo} benchmarks embodied agents in exploration and multi-task domains. Contrasting to these benchmarks, \ourshort distinguishes itself by incorporating a variety of task classes and an array of generalization categories and primarily focuses on evaluating agents' visual generalization abilities.

\paragraph{Generalization.}
How to endow models' generalization abilities is a pivotal topic in machine learning. In computer vision, 
well-established benchmarks are available for exploring distribution shifts and generalization problems~\cite{wiles2021fine,wenzel2022assaying,ye2022ood,koh2021wilds}. While several approaches have been proposed in RL and robotics to tackle such issues~\cite{hansen2021stabilizing,yuan2022pre,huang2022spectrum,bertoin2022look,zhao2022what,kirk2021survey,kirk2023survey,kim2021goal,yang2022towards,xie2023decomposing,yuan2022don}, the benchmarks in use are relatively immature~\cite{elsayed2020ultra,cobbe2020leveraging,stone2021distracting,hansen2021softda}, fraught with numerous limitations, and lacking a unified framework for comparison. For example, Procgen~\cite{cobbe2020leveraging} is a widely used benchmark for quantifying the agents' generalization abilities. However, Procgen remains a video game platform, with 
the human-imaged world rather real-world counterparts, offering limited assistance for agents' generalization in real-world scenarios. The Distracting Control Suite~\cite{stone2021distracting} and DMC-GB~\cite{hansen2021generalization}, building upon DM-Control~\cite{tunyasuvunakool2020dm_control}, introduce some types of visual distractions. Nevertheless, their tasks are solely focused on locomotion, and there remains a substantial gap between this simulated environment and real-world scenarios. By contrast, \ourshort encompasses various forms of generalization, exhibits a high degree of realism, and includes a diverse range of tasks. Avalon~\cite{albrecht2022avalon} is another valuable benchmark for RL generalization. It shares a unified world dynamics and task structure, making it highly suitable as a benchmark for in-distribution generalization. Contrary to Avalon, which is mainly concerned with task-level generalization, \ourshort mainly focuses on out-of-distribution generalization, with a specific concentration on the visual aspects of generalization. 

\section{Conclusion, Limitations, and Future Work}
In this work, we propose a novel Reinforcement Learning benchmark for Visual Generalization~(\ourshort), a comprehensive benchmark for evaluating the visual generalization abilities of trained agents. \ourshort stands apart from existing benchmarks by boasting a broader diversity of tasks and generalization categories, which in turn fosters more persuasive conclusions. According to the quantitative experimental results from \ourshort, we note that, as of now, there are no existing generalization algorithms that can adeptly manage all tasks and generalization types. It is our expectation that the advent of \ourshort will bring fresh perspectives to the research community, and stimulate the advancement of agents that can truly exhibit overall visual generalization capabilities.

\textbf{Limitations.} The agents trained through \ourshort have not yet been evaluated in real-world scenarios. In our future work, we would like to build certain real-world tasks to demonstrate the value that \ourshort can provide in developing generalizable agents for real-world applications.

\section{Acknowledgements}
This work is supported by research program 2022ZD0161700. 

\clearpage

\bibliographystyle{plainnat}
\bibliography{nips2023}


\newpage
\appendix
\section*{Appendix}

\section{Visual Reinforcement Learning Baselines}\label{append:baseline}

\paragraph{DrQ:} This model-free, off-policy reinforcement learning algorithm, is based on Soft Actor-Critic (SAC)~\cite{haarnoja2018soft}. DrQ enhances training stability via applying data augmentation to regularize the Q value of state-action pairs. The key of DrQ is to promote similarity between augmented state-action pairs. The Q-regularization technique is shown in Eq~\ref{eq:DrQ}, where $K$ is the number of samples, $\mathcal{T}$ is the collection of augmentation.
\begin{equation}\label{eq:DrQ}
\mathbb{E}_{\substack{s \sim \mu(\cdot) \\ a \sim \pi(\cdot \mid s)}}[Q(s, a)] \approx \frac{1}{K} \sum_{k=1}^K Q\left(f\left(s^*, \nu_k\right), a_k\right) \text { where } \nu_k \in \mathcal{T} \text { and } a_k \sim \pi\left(\cdot \mid f\left(s^*, \nu_k\right)\right)
\end{equation}

\paragraph{DrQ-v2:} An improved version of DrQ. DrQ-v2 fuses essential elements from the DDPG algorithm with data augmentation to strengthen visual RL agents' performance. DrQ-v2 also incorporates techniques such as n-step return and target critic, leading to commendable results in most of the medium and hard level DM-Control tasks. The TD-target is defined as follows, where $x_{t+n}$ is the n-step observation, $\boldsymbol{a}_{t+n}$ is the n-step action, and $\bar{\theta}_{1,2}$ is the Q-target networks:

\begin{equation}
y=\sum_{i=0}^{n-1} \gamma^i r_{t+i}+\gamma^n \min _{k=1,2} Q_{\bar{\theta}_k}\left({aug(x_{t+n})}, \boldsymbol{a}_{t+n}\right)
\end{equation}

\paragraph{CURL:} CURL integrates contrastive learning methods into the reinforcement learning training process. The auxiliary contrastive loss~(Eq~\ref{eq:curl}) allows the agent to obtain better image representation during training, thus mitigating the optimization difficulty under high-dimensional inputs. In our implementation, we only apply a single encoder to produce visual representations instead of two polyak-averaging encoders. This alteration improves the sample efficiency of CURL and put it on a comparable performance with DrQ-v2. More experiments are shown in Appendix~\ref{appendix:re-implement}.

\begin{equation}\label{eq:curl}
\mathcal{L}_q=\log \frac{\exp \left(q^T W k_{+}\right)}{\exp \left(q^T W k_{+}\right)+\sum_{i=0}^{K-1} \exp \left(q^T W k_i\right)}
\end{equation}

\paragraph{PIE-G:}
PIE-G proposes a simple yet effective method, combining ImagNet pre-trained visual representations with the early layer and updates of BatchNorm statistical parameters to further enhance the generalization ability of the agent.

\paragraph{SVEA:} SVEA finds that heavy data augmentation introduces additional high variance to agent training, which can lead to instability or even divergence. SVEA suggests that using the Q-values of non-augmented images as the target of estimated Q-values for augmented images (Eq~\ref{eq:svea}), thus stabilizing the variance of the value estimation. 

\begin{equation}\label{eq:svea}
\left\|Q_\theta\left(aug(x_{t}), \mathbf{a}_t\right)-q_t^{\mathrm{tgt}}\right\|_2^2
\end{equation}

\paragraph{SRM:}
SRM proposes a novel data augmentation method that operates in the frequency domain. It helps diversify data and alleviate distribution shift issues under various visual scenarios. During the training stage, SRM randomly discards parts of the frequency information from observations, forcing the policy to select suitable actions based on the remaining information. The augmentation method is shown in Eq~\ref{eq:srm},  where $\mathcal{F}$ is the fast Fourier transform, $\mathbf{M}$ is a binary masking matrix, and $\mathbf{Z}$ is a random noise image.

\begin{equation}
\hat{\mathcal{F}}\left(o_i\right)=\mathbf{M} \cdot \mathcal{F}\left(o_i\right)+(\mathbf{1}-\mathbf{M}) \cdot \mathcal{F}(\mathbf{Z})
\label{eq:srm}
\end{equation}

\paragraph{SGQN:}
This algorithm introduces the saliency map for the use of augmenting images. Saliency maps, a tool used in computer vision, offers an interpretability analysis of encoders. SGQN retains only agent's focusing areas and removes the visual background by the generated saliency map. This approach utilizes the augmentation objectives in SVEA~\cite{hansen2021stabilizing} to further improve the model's generalization performance. The auxiliary objective is shown in Eq~\ref{eq:sgqn}, where $M_\rho((o, a), a)$ is the binary masking matrix introduced from the saliency map.

\begin{equation}\label{eq:sgqn}
L_C(\theta)=\| Q_\theta\left(o, a\right)-Q_\theta\left(o \odot M_\rho(o, a), a\right) \|^2
\end{equation}

\section{Implementation Details}\label{append:details}

\paragraph{Indoor navigation:} Habitat serves as the simulator and extends a variety of indoor navigation tasks. We select \emph{ImageNav} as the test env, whose goal is defined by the image of target location in the chosen map. Due to the complexity in the default training and validation episode settings, which demands extensive training periods to achieve a satisfactory standard, we simplify the setup to 500 initial positions and 1 target position. Meanwhile, we utilize the 3D scenes from the Gibson dataset as our map for all experiments.

\paragraph{Autonomous driving:} We choose the stable version of CARLA 0.9.10 for simulation. The reward function is adopted 
 from Zhang et al.~\cite{zhang2020learning}. We also implement the wrapping methods from Huang et al.~\cite{CarlaEnv} for novel CARLA environments. Moreover, to enhance exploration and ensure stable training, we standardized the \emph{std\_schedule} across all algorithms. Each difficulty level contains two weathers, \emph{Easy level:} soft\_high\_light, soft\_noisy\_low\_light; \emph{Medium level:} HardRainSunset, SoftRainSunset; \emph{Hard level:} hard\_low\_light, hard\_noisy\_low\_light. The aggregated return is calculated by averaging over the weather at the same level. Further details can be accessed in the documentation provided within our GitHub repository.

\paragraph{Dexterous manipulation:} In \ourshort, we select three single-view Adroit tasks. Given that tasks in Adroit typically necessitate demonstrations for successful completion, we employ VRL3, the state-of-the-art baseline for these tasks. Since the update process of VRL3 is based on DrQ-v2, it allows a seamless transfer of our algorithms to VRL3's codebase. There are three stages for VRL3 training: stage1 responses to gain a basic perception ability via pretraining on ImageNet; stage2 utilizes offline RL training with expert demonstrations; stage3 executes online training. It is noteworthy to mention that the experiment of VRL3 demonstrates that in single view tasks, only applying stage3 is sufficient to accomplish Adroit tasks with high sample efficiency. Therefore, to compare each algorithm more effectively, we exclude the use of stage1 and stage2. The aggregated success rate is calculated by averaging over all three tasks.

\paragraph{Table-top manipulation:} SECANT~\cite{pmlr-v139-fan21c} previously employed Robosuite for generalization testing. Building on its codebase, we adopt one of the latest versions - Robosuite 1.4.0 and mujoco 2.3.0 as well as simplified the installation process. Meanwhile, we also introduce a range of new classes of visual generalization. For each difficulty level, we deploy a variety of scenarios, and each trained agent is evaluated within each environment 10 times~(in a total of 100 evaluations). The aggregated return is calculated by averaging over all three tasks.

\paragraph{Locomotion:} In addition to the locomotion tasks from DM-Control~(1.0.8 version), we also incorporate models from Mujocoreie~\cite{menagerie2022github}, and carefully designed corresponding rewards, enabling them to accomplish \textit{walk} or \textit{stand} tasks. Furthermore, building on DMC-GB, we have added additional generalization categories for further enriching \ourshort. The aggregated return is calculated by averaging over two tasks.

Our experiments are all conducted with TeslaA40 or TeslaA100 GPU and AMD EPYC 7542 32-Core Processor CPU. More details can be founded in \url{https://github.com/gemcollector/RL-ViGen}. 

\section{Hyper-parameters}\label{append:hyper}
We use the same hyper-parameters as the original papers and perform a small-scale grid search to achieve better performance of certain algorithms. The common hyper-parameters are listed in Table~\ref{table: common_hyper}.

\begin{table}[h]
\centering
\caption{Common hyper-parameters in \ourshort.}
\renewcommand\tabcolsep{24.0pt}
\begin{tabular}{ccccc}
\toprule[0.5mm]
Hyper-parameters                             & Value                                                                   \\ \hline
Input size                            & 84 $\times$ 84   \\
Discount factor $\gamma$               & 0.99            \\
Replay Buffer size                      & int(1e7)            \\
Feature dim         &                 DrQ(v2), CURL: 50, otherwise: 256 \\
Action repeat & Robosuite: 1, otherwise: 2 \\
N-step return  &  DrQ: 1, otherwise: 3 \\
Optimizer & Adam \\
Hidden dim & 1024 \\
Frame stack & 3  \\

\hline
\end{tabular}
\label{table: common_hyper}
\end{table}

The individual hyper-parameters are listed in the following Tables. The additional hyper-parameters introduced by SGQN are listed as well.

\begin{table}[h]
\centering
\caption{CARLA hyper-parameters in \ourshort.}
\renewcommand\tabcolsep{24.0pt}
\begin{tabular}{ccccc}
\toprule[0.5mm]
Hyper-parameters                             & Value                                                                   \\ \hline
Training Frames                            & int(1e6)   \\
Learning Rate  &     PIE-G: 5e-5, DrQ: 5e-4, otherwise: 1e-4 \\
N-step return &  1 \\
SGQN quantile & 0.9 \\
SGQN critic weight & 0.5 \\
SGQN aux lr & 8e-5 \\
\hline
\end{tabular}
\label{table: carla_hyper}
\end{table}
\begin{table}[h]
\centering
\caption{Habitat hyper-parameters in \ourshort.}
\renewcommand\tabcolsep{48.0pt}
\begin{tabular}{ccccc}
\toprule[0.5mm]
Hyper-parameters                             & Value                                                                   \\ \hline
Training Frames                            & int(7e5)   \\
Learning Rate  &     1e-4 \\
N-step return &  1 \\
SGQN quantile & 0.93 \\
SGQN critic weight & 0.9 \\
SGQN aux lr & 8e-5 \\
\hline
\end{tabular}
\label{table: habi_hyper}
\end{table}
\begin{table}[h]
\centering
\caption{Adroit hyper-parameters in \ourshort.}
\renewcommand\tabcolsep{24.0pt}
\begin{tabular}{ccc}
\toprule[0.5mm]
Hyper-parameters  & Task & Value \\ 
\midrule
\multirow{3}{*}{Training Frames} & Hammer & int(1e6) \\ 
& Door & int(1e6) \\ 
& Pen & int(2e6) \\ 
\midrule
\multirow{3}{*}{Learning Rate} & Hammer & 1e-4 \\ 
& Door & 1e-4  \\ 
& Pen & 1e-4 \\ 
\midrule
\multirow{3}{*}{SGQN quantile } & Hammer & 0.9 \\ 
& Door & 0.9 \\ 
& Pen & 0.9 \\ 
\midrule
\multirow{3}{*}{SGQN critic weight } & Hammer & 0.9 \\ 
& Door & 0.5 \\ 
& Pen & 0.9 \\ 
\midrule
\multirow{3}{*}{SGQN aux lr } & Hammer & 8e-5 \\ 
& Door & 8e-5 \\ 
& Pen & 8e-5 \\ 
\bottomrule[0.5mm]
\end{tabular}
\label{table: adroit_hyper}
\end{table}
\begin{table}[h]
\centering
\caption{Robosuite hyper-parameters in \ourshort.}
\renewcommand\tabcolsep{12.0pt}
\begin{tabular}{ccc}
\toprule[0.5mm]
Hyper-parameters  & Task & Value \\ 
\midrule
\multirow{3}{*}{Training Frames} & Door & int(6e5) \\ 
& Lift & int(8e5) \\ 
& TwoArmPegInhole & int(8e5) \\ 
\midrule
\multirow{3}{*}{Learning Rate} & Door & 1e-4 \\ 
& Lift & DrQ(v2), CURL: 1e-4, otherwise: 8e-5  \\ 
& TwoArmPegInhole & SGQN: 1e-4, otherise: 8e-5 \\ 
\midrule
\multirow{3}{*}{Level} & Door & Easy \\ 
& Lift & Medium \\ 
& TwoArmPegInhole & Medium \\ 
\midrule
\multirow{3}{*}{SGQN quantile } & Door & 0.9 \\ 
& Lift & 0.9 \\ 
& TwoArmPegInhole & 0.87 \\ 
\midrule
\multirow{3}{*}{SGQN critic weight } & Door & 0.7 \\ 
& Lift & 0.7 \\ 
& TwoArmPegInhole & 0.7 \\ 
\midrule
\multirow{3}{*}{SGQN aux lr } & Door & 8e-5 \\ 
& Lift & 8e-5 \\ 
& TwoArmPegInhole & 8e-5 \\ 
\bottomrule[0.5mm]
\end{tabular}
\label{table: robo_hyper}
\end{table}

\section{Visualization of each difficulty level}\label{append:diff_level}
To gain a better understanding of our setting and \ourshort, we visualize the images under various generalization settings and difficulty levels as mentioned in the experiment section.

\subsection{Visual Appearances and Lighting Changes}\label{append:visualization}

\subsubsection{Robosuite}
In the context of Robosuite, each difficulty level comprises $10$ distinct scenes. We perform $10$ trials for each of these scenes~(100 trials in total). 

\begin{figure}[h]
  \centering
  \includegraphics[width=1.0\linewidth]{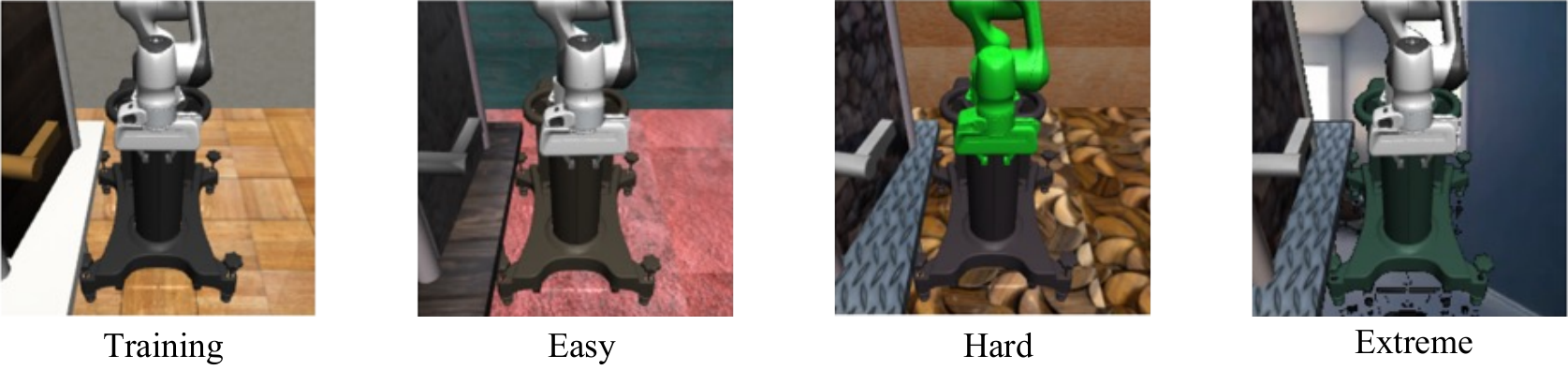}
  \caption{\textbf{The visualization of various difficulty levels of Robosuite.} This figure shows examples from the \textit{Door} task. As the difficulty level increases, more types of distracting factors are introduced.}
  \label{fig:robo_app}
\end{figure}

The \textit{Easy} level includes changes of the background appearance, while the \textit{Hard} level contains additional complexities of moving light and alterations to the robotic arm's color. The \textit{Extreme} level further employs a dynamic video background to evaluate the trained agents' generalization abilities. The visualized figures are shown in Figure~\ref{fig:robo_app}. 

\subsubsection{CARLA}

\begin{figure}[h]
  \centering
  \includegraphics[width=1.0\linewidth]{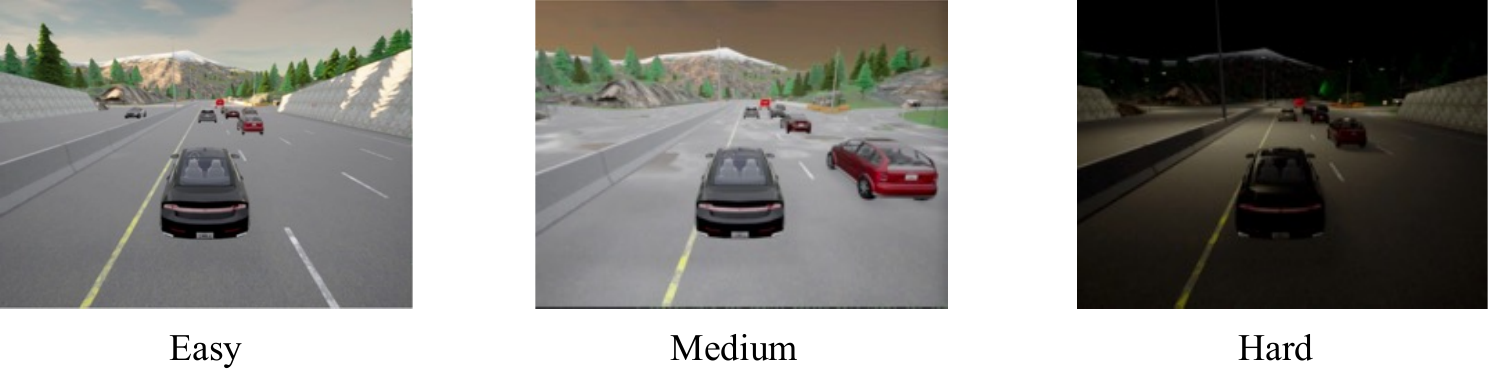}
  \caption{\textbf{The visualization of various difficult level of CARLA.} The higher the disparity from the training observations, the more challenging the new scenario. }
  \label{fig:carla_app}
\end{figure}

Apart from the default weather settings in CARLA, we implement a series of challenging new scenarios. In our CARLA setup, each level of difficulty is characterized by two specific weather conditions. The \textit{Easy} level includes \textit{soft\_noisy\_low\_light} and \textit{soft\_high\_light}, while the \textit{Medium} level is defined by the \textit{HardRainSunset} and \textit{SoftRainSunset} conditions. The \textit{Hard} level contains \textit{hard\_low\_light} and \textit{hard\_noisy\_low\_light}. As the disparity between the novel scenarios and the training images increases, the difficulty level of generalization grows. \ourshort also encompasses challenging conditions such as rainy, overcast, and slippery road surfaces. The visualized figures are shown in Figure~\ref{fig:carla_app}.

\subsubsection{Habitat}
\begin{figure}[h]
  \centering
  \includegraphics[width=1.0\linewidth]{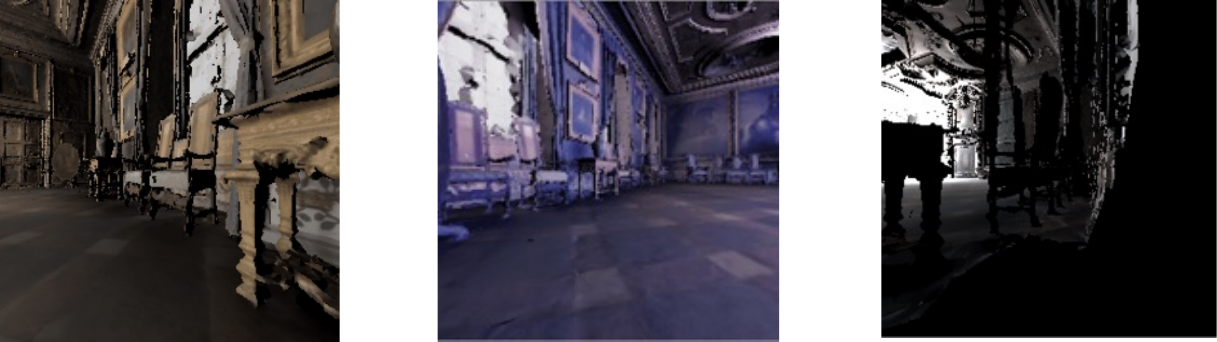}
  \caption{\textbf{The visualization of Habitat.} We create 10 distinct scenarios for the generalization of visual appearances.}
  \label{fig:habi_app}
\end{figure}

For Habitat, the Gestaltor 3D model editor is applied to modify the appearance of the scene's 3D models. A total of $10$ distinct scenarios are created. The visualized figures are shown in Figure~\ref{fig:habi_app}.

\subsubsection{Locomotion}
\begin{figure}[h]
  \centering
  \includegraphics[width=0.9\linewidth]{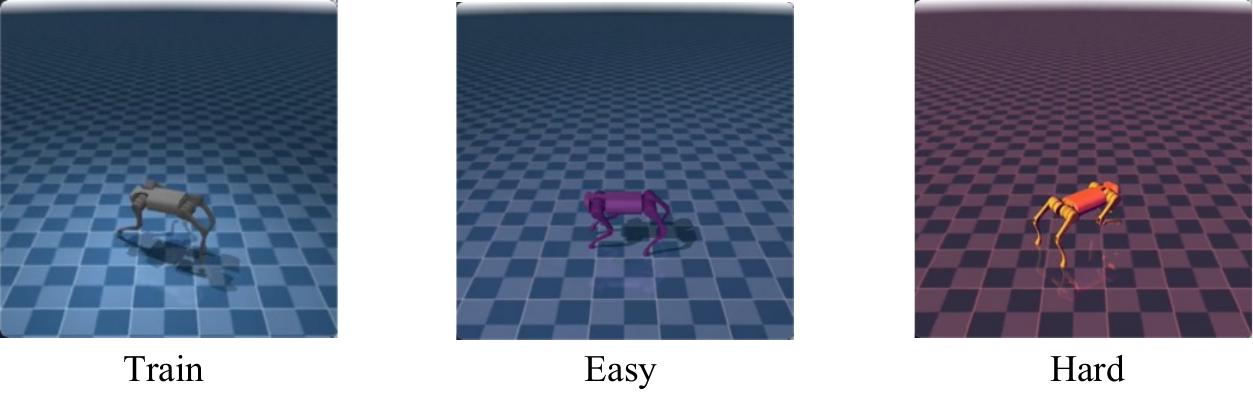}
  \caption{\textbf{The visualization of various difficulty level of DM-Control.} The figure above show examples from unitree tasks. Factors such as light color, light position, movement of light, and object color are varied.}
  \label{fig:loco_app}
\end{figure}

For DM-Control, we further augment numerous new types of generalizations on the basis of DMC-GB. For the unitree series tasks, \textit{Easy} and \textit{Hard} denote two levels of difficulty regarding light color, light position, changes of light's movement and objects' color. The visualized figures are shown in Figure~\ref{fig:loco_app}.

\subsubsection{Adroit}
\begin{figure}[h]
  \centering
  \includegraphics[width=0.9\linewidth]{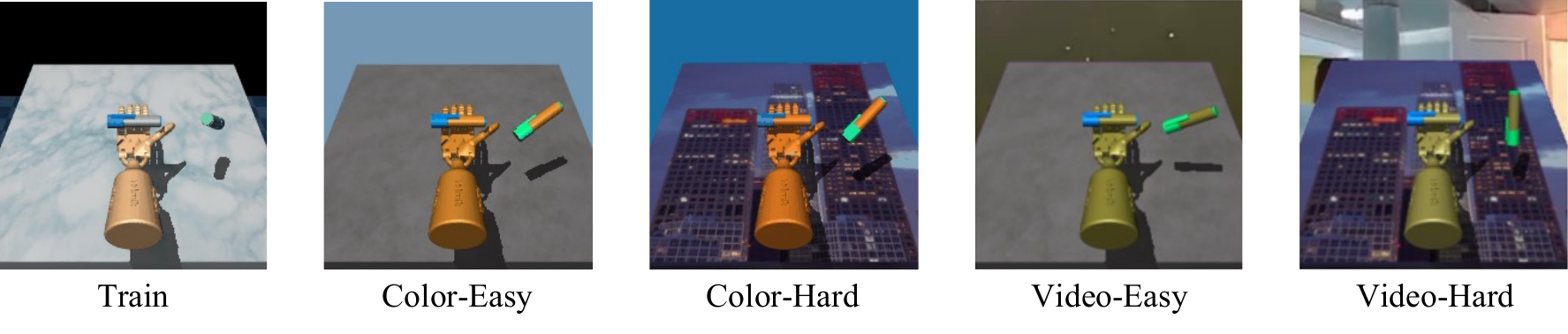}
  \caption{\textbf{The visualization of various difficulty levels of Adroit.} This figure demonstrates examples from the \textit{Pen} task. We show four generalization scenarios provided in \ourshort. }
  \label{fig:adroit_app}
\end{figure}

In the Adroit environment, we provide four generalization scenarios. The \textit{Color} setting changes the background, object color, and table texture, while the \textit{Video} setting utilizes a dynamic background and introduces moving light. As illustrated in Figure~\ref{fig:adroit_app}, each scenario is configured with two levels of difficulty. 

\subsection{Camera Views}
\begin{figure}[h]
  \centering
  \includegraphics[width=0.8\linewidth]{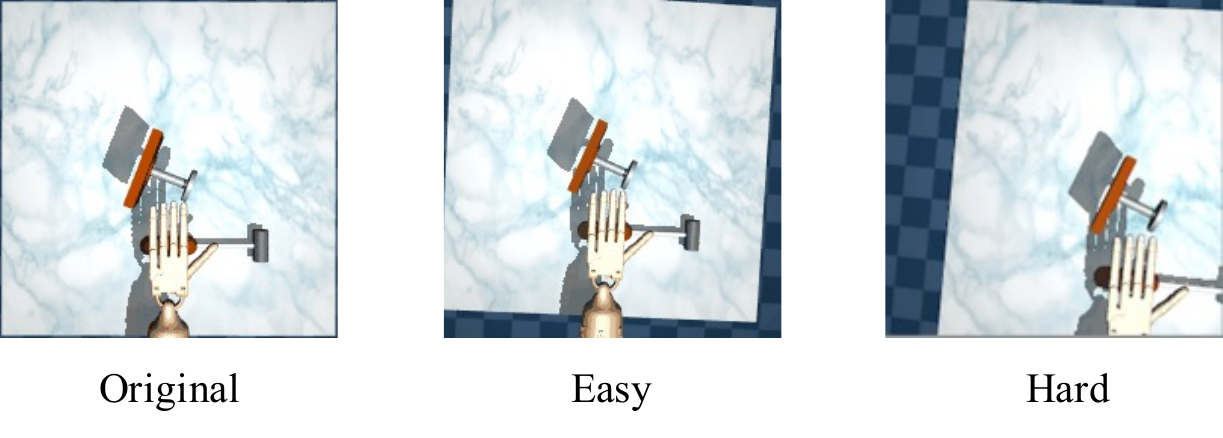}
  \caption{\textbf{The visualization of camera views of Adroit.} The larger the deviation angle of the camera, the higher the difficulty of generalization.}
  \label{fig:adroit_cam}
\end{figure}

For camera view generalization, we implement alternations to the camera view through the modification of camera's orientation, position, and FOV. The visualized figure is shown in Figure~\ref{fig:adroit_cam}.

\subsection{Scene Structures}
\begin{figure}[h]
  \centering
  \includegraphics[width=0.9\linewidth]{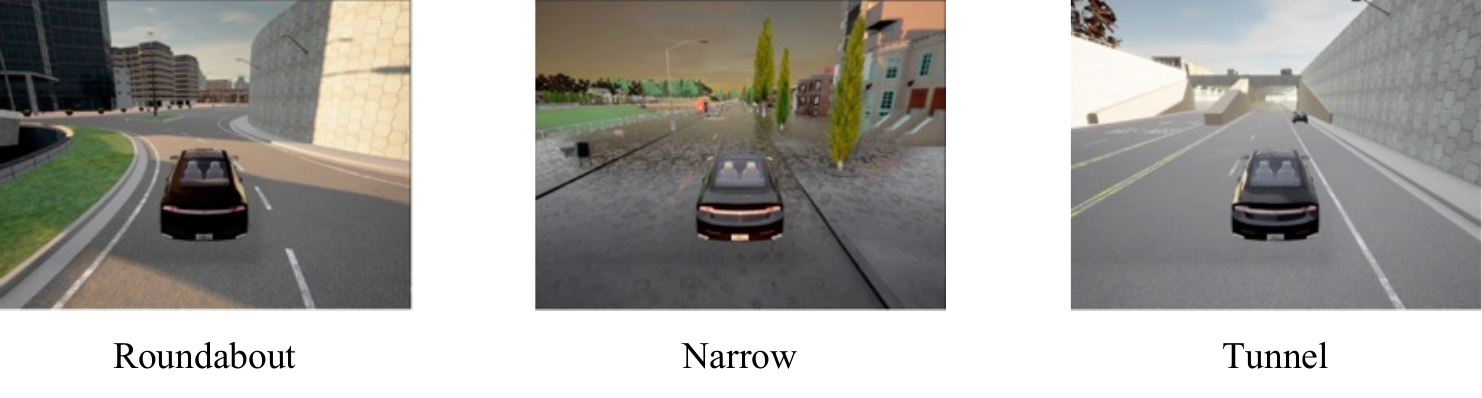}
  \caption{\textbf{The visualization of scene structures of CARLA.} 
We selected certain locations within different maps to serve as scenarios for scene structure generalization.}
  \label{fig:carla_scene}
\end{figure}

As shown in Figure~\ref{fig:carla_scene}, we established a variety of road scenarios in CARLA, including roundabouts, narrow paths, tunnels, etc., which can be also utilized in conjunction with other adjustable parameters. As the experiment illustrated in Figure~\ref{fig:scene_score}, we employ the same weather conditions as those during training.

\subsection{Cross Embodiments}
In terms of cross-embodiment generalization, we modify the type of the robotic arm in Robosuite. In addtion, by leveraging the OSC\_POSE control method, the input actions are interpreted as delta values from the current state, thus facilitating to maintain the action space dimensions and corresponding meanings. 

\begin{figure}[h]
  \centering
  \includegraphics[width=0.8\linewidth]{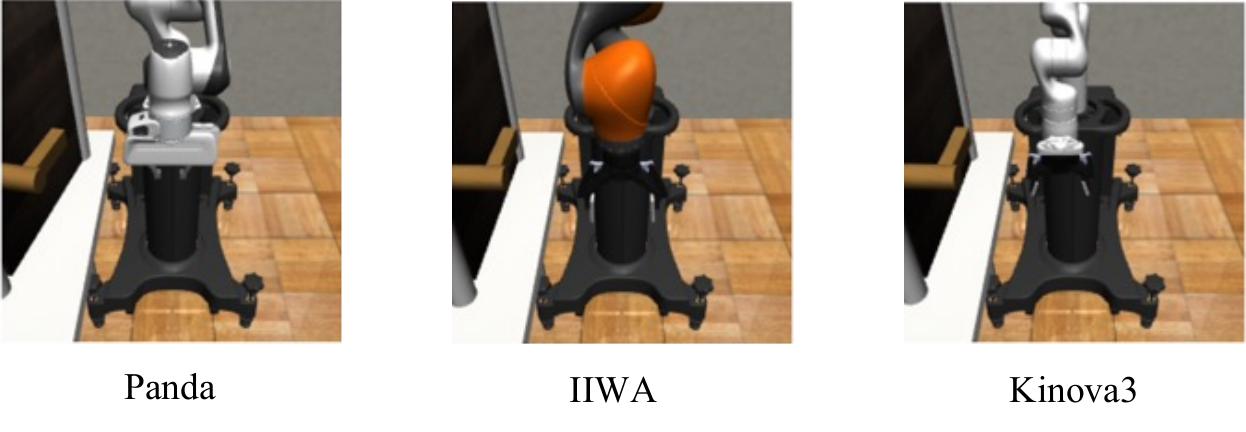}
  \caption{\textbf{The visualization of cross embodiment of Robosuite.} This figure shows examples from the \textit{Door} task. Here, we demonstrate our modification of the style of the robotic arm for cross-embodiment generalization. }
  \label{fig:robo_cross}
\end{figure}

\section{Additional Results}\label{append:add_results}

\subsection{Generalization Evaluation}

\subsubsection{Locomotion}

\begin{figure}[h]
  \centering
  \includegraphics[width=0.8\linewidth]{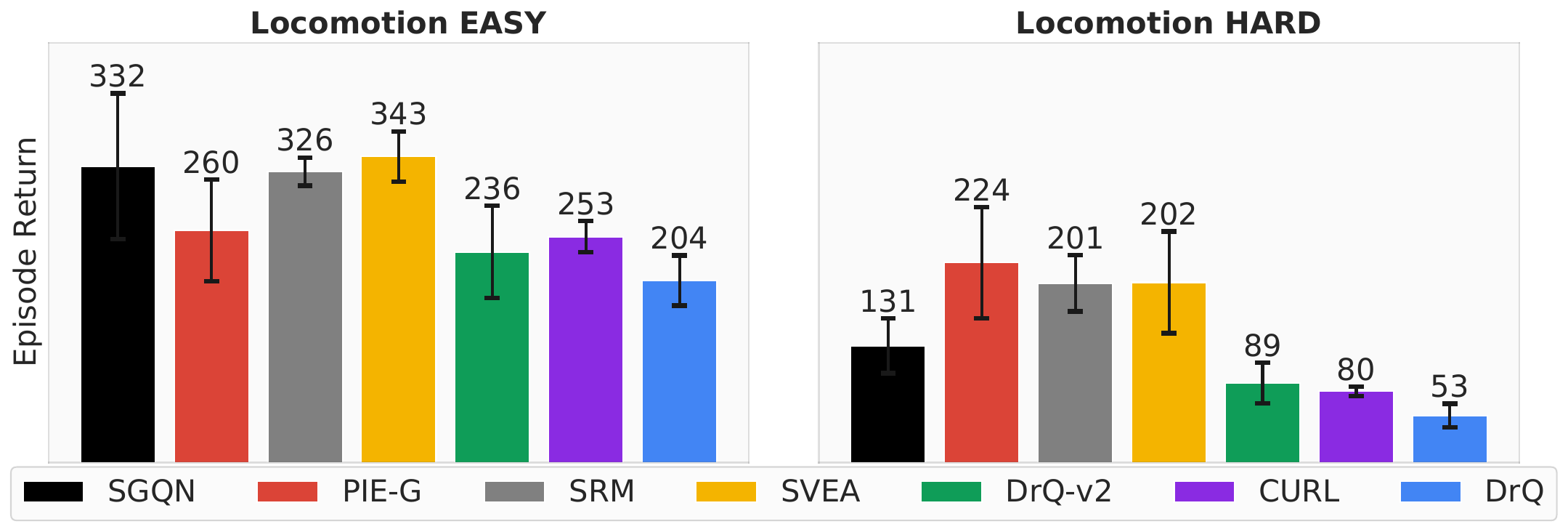}
  \caption{\textbf{Generalization score of Locomotion.} The generalization algorithms show comparable performance at two difficulty levels. }
  \label{fig:loco_score}
\end{figure}

Built upon DM-Control, which has included numerous locomotion tasks, we extend this benchmark by integrating real-world robot models from Mujocoreie~\cite{menagerie2022github} with corresponding tasks. Moreover, \ourshort also augments DMC-GB with more tasks and generalization types. Here we evaluate the performance of each algorithm on the Unitree series tasks. Figure~\ref{fig:loco_score} demonstrates that all generalization algorithms exhibit comparable performance. More specifically, SVEA outperforms other techniques in the \textit{Easy} setting, where the other generalization techniques do not show any advantages. In the \textit{Hard} setting, where the agent's color closely resembles that of the surrounding environment, SGQN may not effectively capture the agent's outline, leading to a performance decline.

\subsubsection{Table-top Manipulation}
\begin{figure}[h]
  \centering
  \includegraphics[width=0.9\linewidth]{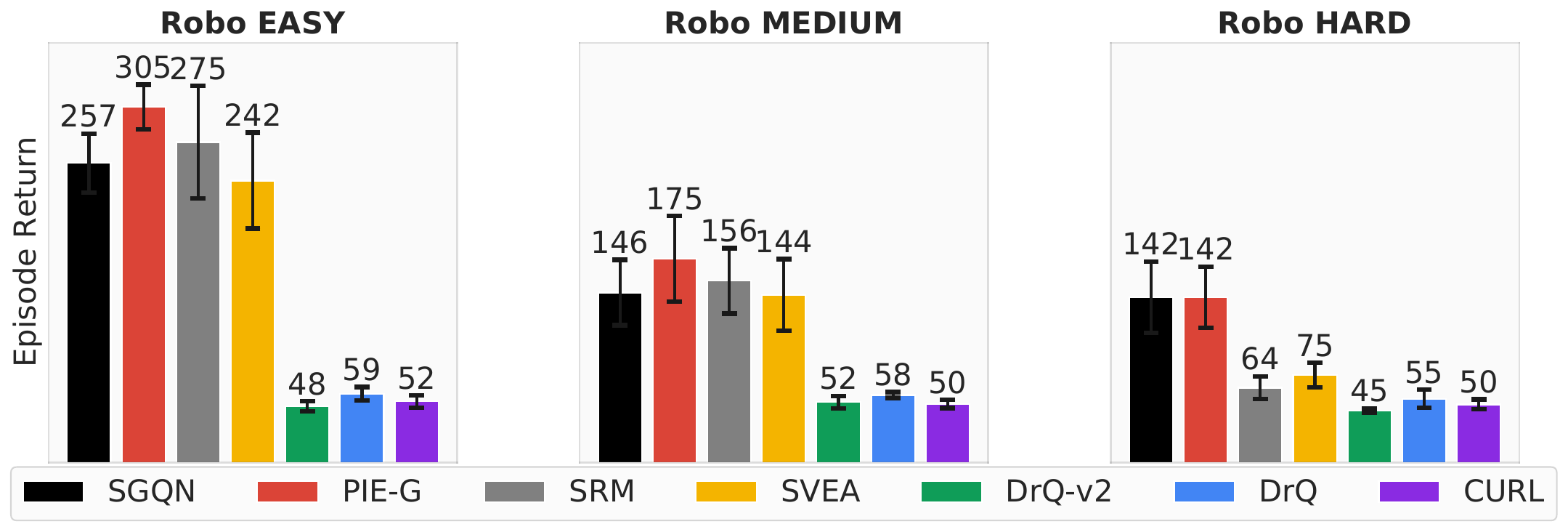}
  \caption{\textbf{The aggregated generalization score of table-top manipulation.} We present the aggregated return of three tasks for each method. PIE-G shows better generalization performance of table-top manipulation tasks when facing unseen visual scenarios.}
  \label{fig:robo_score}
\end{figure}

In Robosuite, three tasks, including single-arm and dual-arm settings, are selected in \ourshort: \emph{Door}, \emph{Lift}, and \emph{TwoArmPegInhole}. Additionally, we create multiple difficulty levels, incorporating various visual scenarios, and dynamic backgrounds. In the \emph{Easy} and \emph{Medium} test environments, where considerable variations in visual colors and lighting changes are introduced,  the results in Figure~\ref{fig:robo_score} show that PIE-G demonstrates slightly better performance than that of SGQN and SRM in \emph{Easy} and \emph{Medium} settings. However, when faced with the \emph{Hard} setting that integrates dynamic video backgrounds, SRM, which mainly resorts to static frequency-based augmentation, is unable to adapt effectively to such scenarios for completing the manipulation tasks. Figure~\ref{fig:robo_score} further indicates that the remaining algorithms struggle to demonstrate generalization abilities in this environment.

\subsection{Wall Time}

\begin{wrapfigure}[16]{r}{0.4\textwidth}%
    \centering
    \includegraphics[width=0.4\textwidth]{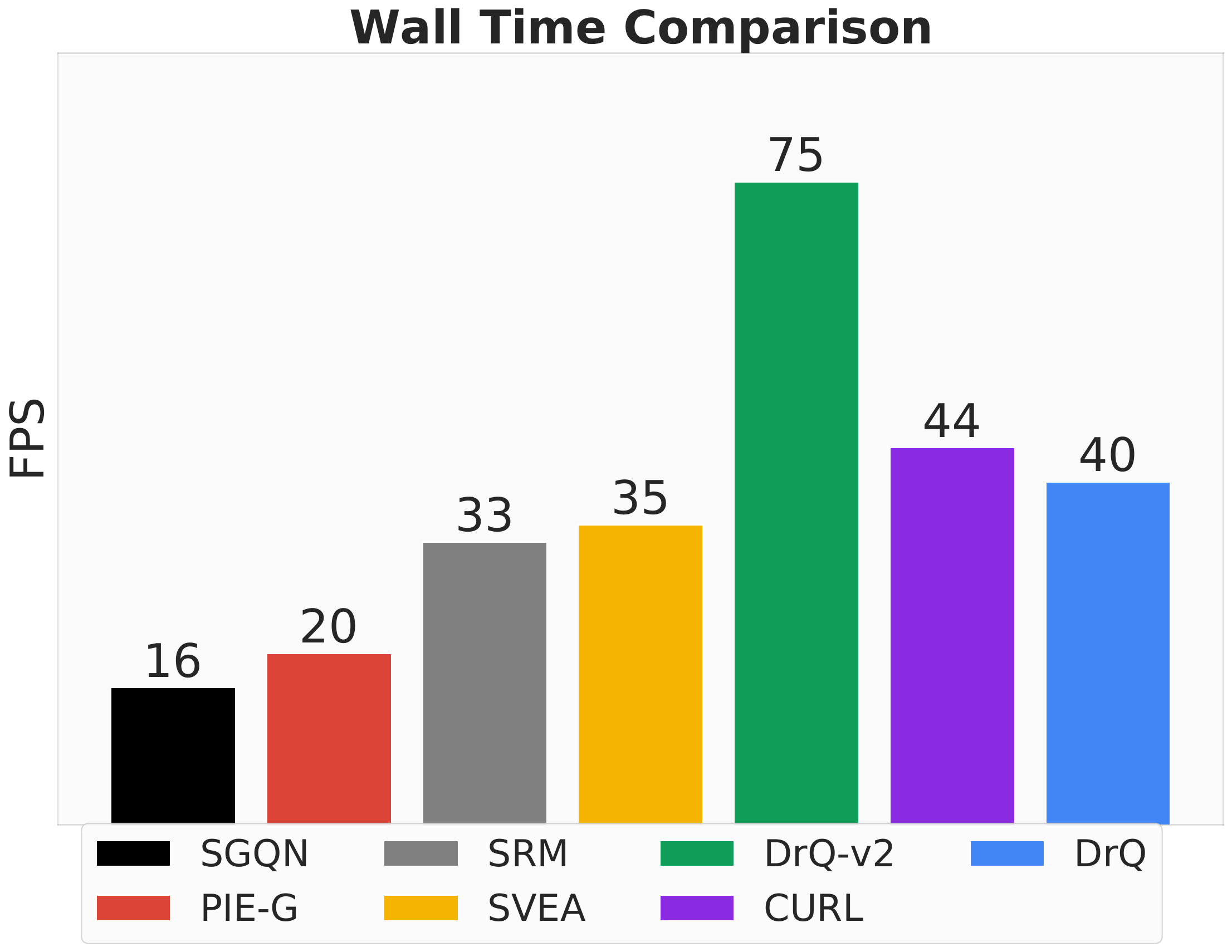}
    \caption{\textbf{Wall Time Comparison.} DrQ-v2 enjoys the lowest computational cost.}
    \label{fig:wall_time}
\end{wrapfigure}
So far, our main focus has been the comparison of generalization performance of each method across various tasks. In this section, we turn our attention to the comparison of each algorithm's wall-clock training time. We choose \textit{Walker walk} task from DMControl for evaluation. This task requires a large batch size for training, thus is suitable for better demonstrating the wall-time efficiency of each approach. Frames-per-second~(FPS) is selected to be the evaluation metric. Figure~\ref{fig:wall_time} illustrates that DrQ-v2 owns the least computational cost. Conversely, for the algorithms that utilize additional data for augmentation purposes, they tend to exhibit lower frames-per-second~(FPS) rates. SGQN builds the saliency maps during every training step, which takes extra costs. Meanwhile, PIE-G utilizes the ImageNet pre-trained ResNet model to convert high-dimensional images into representations, thus adding more burden on the model's inference compared to other algorithms.

\begin{figure}[t]
  \centering
  \includegraphics[width=1.0\linewidth]{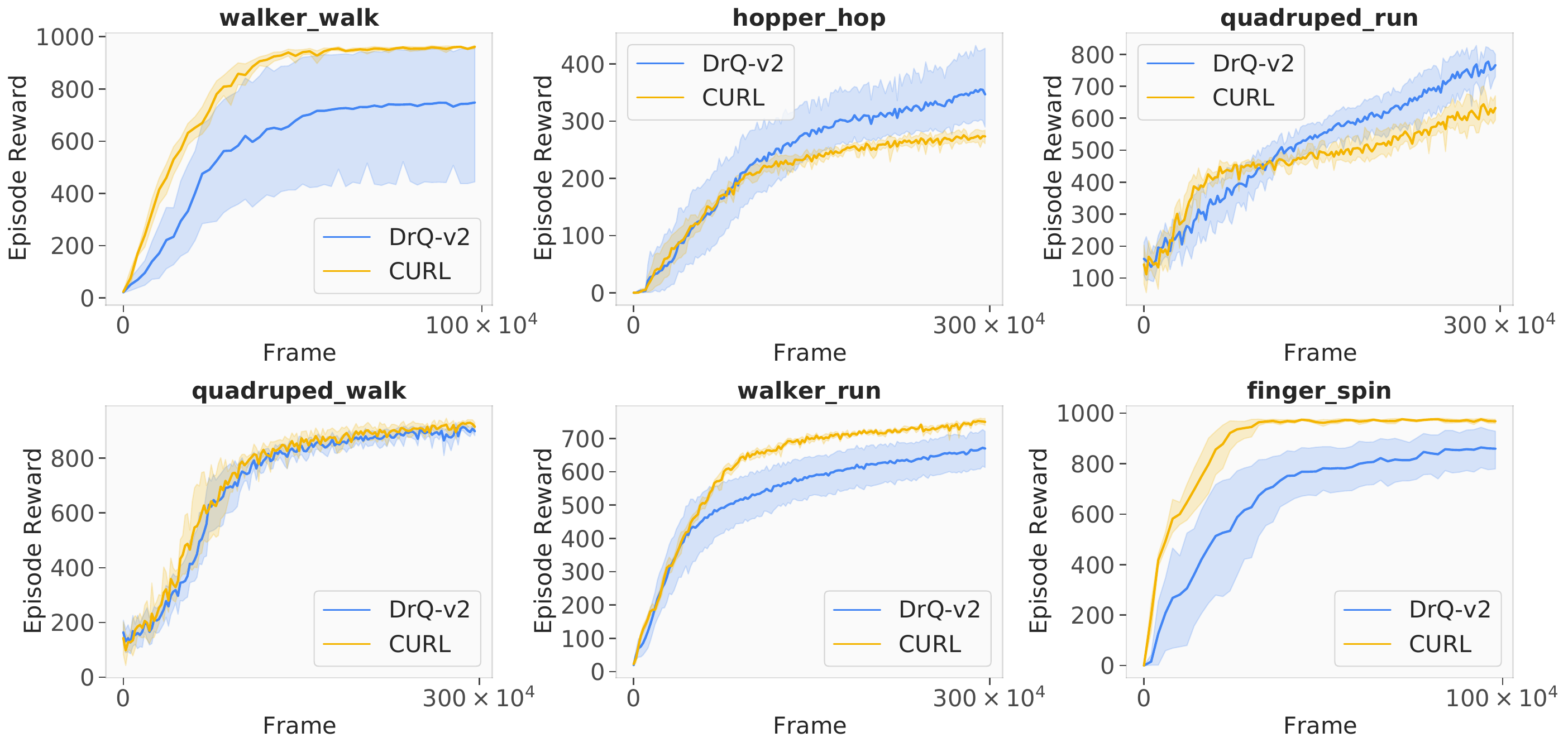}
  \caption{\textbf{The sample efficiency comparison between CURL and DrQ-v2.} Our re-implementation of CURL can achieve comparable sample efficiency with DrQ-v2.}
  \label{fig:curl_curve}
\end{figure}

\subsection{The re-implementation of CURL}\label{appendix:re-implement}
CURL~\cite{laskin2020curl}, which adopts contrastive loss as an auxiliary objective, is frequently mentioned in numerous works~\cite{yarats2021mastering, kostrikov2020image, laskin2020reinforcement}, yet the effectiveness of contrastive loss appears to be less pronounced~\cite{laskin2020reinforcement, li2022does}. Distinct from prior studies, we do not utilize a target encoder and remove the update of momentum parameters related to the encoder. As shown in Figure~\ref{fig:curl_curve}, comparing to the state-of-the-art approach DrQ-v2 and the results reported in previous work~\cite{yarats2021mastering}, the use of a single shared encoder for achieving representations seems to yield more favorable results when leveraging contrastive loss.


\subsection{Sample efficiency}\label{appendix: sample}
\begin{figure}[t]
  \centering
  \includegraphics[width=1.0\linewidth]{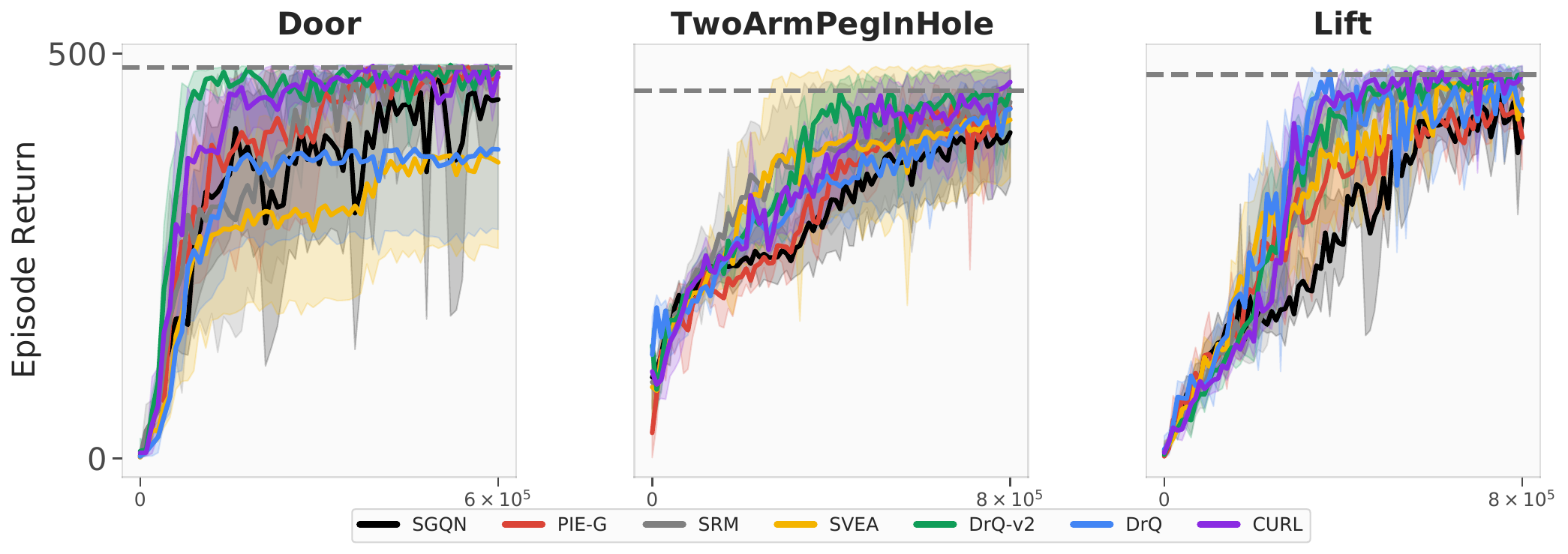}
  \caption{\textbf{Sample efficiency of Robosuite.} The episode return of each algorithm. We normalize the training steps into (0, 1). DrQ-v2 and CURL show better sample efficiency. }
  \label{fig:robo_curve}
\end{figure}

\begin{figure}[t]
    \centering
    \setlength{\abovecaptionskip}{5pt}
    \subfloat[ \label{fig:stand_curve}]
        {\includegraphics[width=0.4\linewidth]{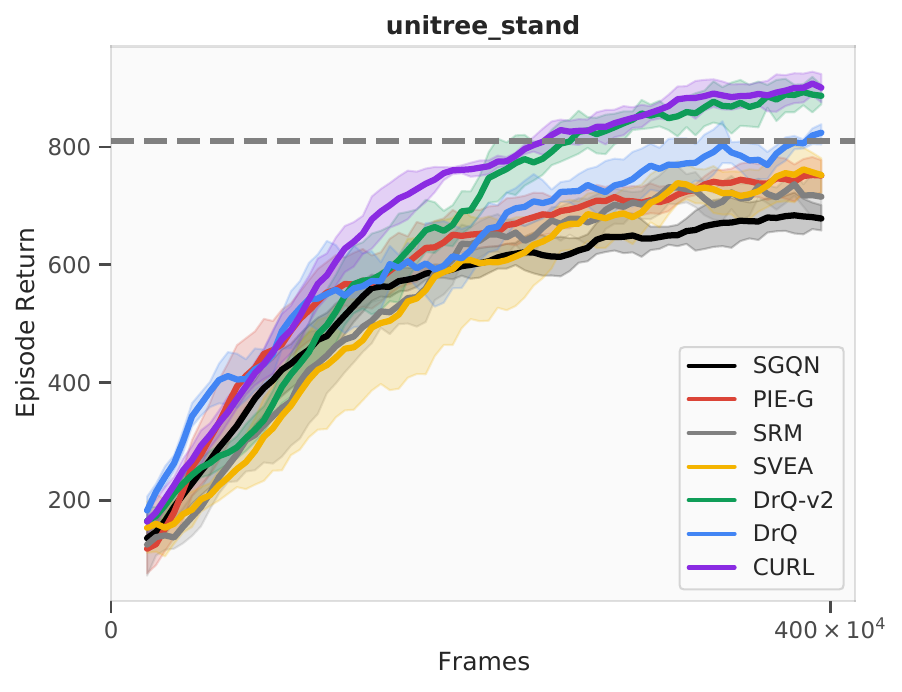}}
    \hspace{6mm}
    \subfloat[ \label{fig:walk_curve}]
        {\includegraphics[width=0.4\linewidth]{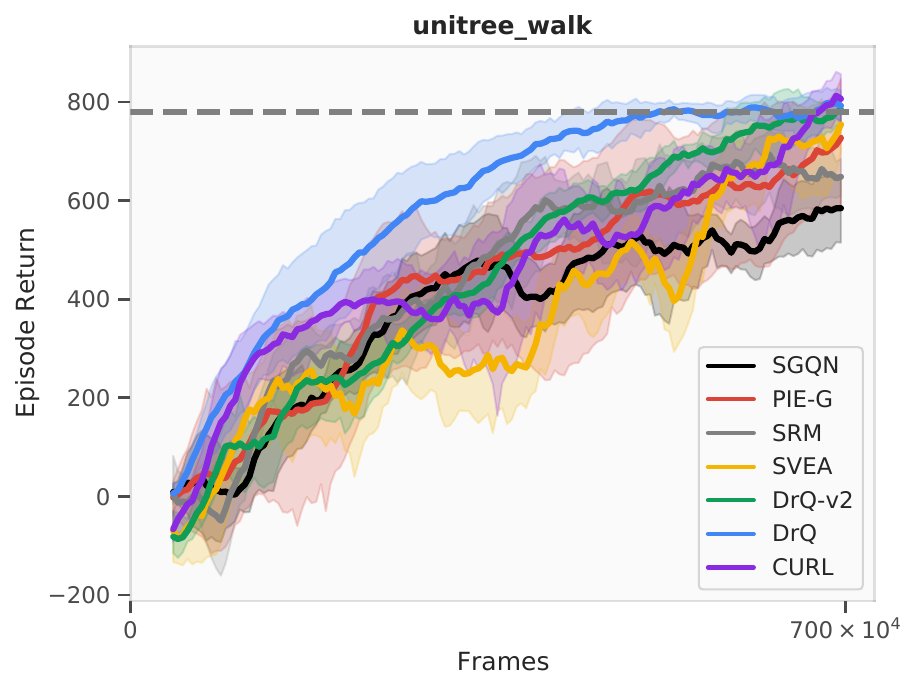}}
\vspace{-5pt}
\caption{\textbf{Sample efficiency of Unitree tasks.} The episode return of each method. The agent, positioned at the center of observation in these tasks, is subjected to additional noise due to data augmentation.}
\label{fig: loco_curve}
\end{figure}

In this section, we compare the sample efficiency of various visual RL algorithms. As one of the state-of-the-art visual RL algorithms, DrQ-v2 serves as a baseline for evaluating the training performance of various algorithms across different tasks. In each figure, the convergence performance of DrQ-v2 is marked with a gray dashed line.As shown in Figure~\ref{fig:robo_curve} and Figure~\ref{fig: loco_curve}, DrQ-v2 and CURL obtain advantageous sample efficiency in locomotion and table-top manipulation tasks. A shared attribute between these two types of tasks is that the agent is positioned at the center of observation. Hence, the additional noise introduced by data augmentation tends to exacerbate training instability.

\begin{figure}[ht]
    \centering
    \setlength{\abovecaptionskip}{5pt}
    \subfloat[ \label{fig:habi_curve}]
        {\includegraphics[width=0.4\linewidth]{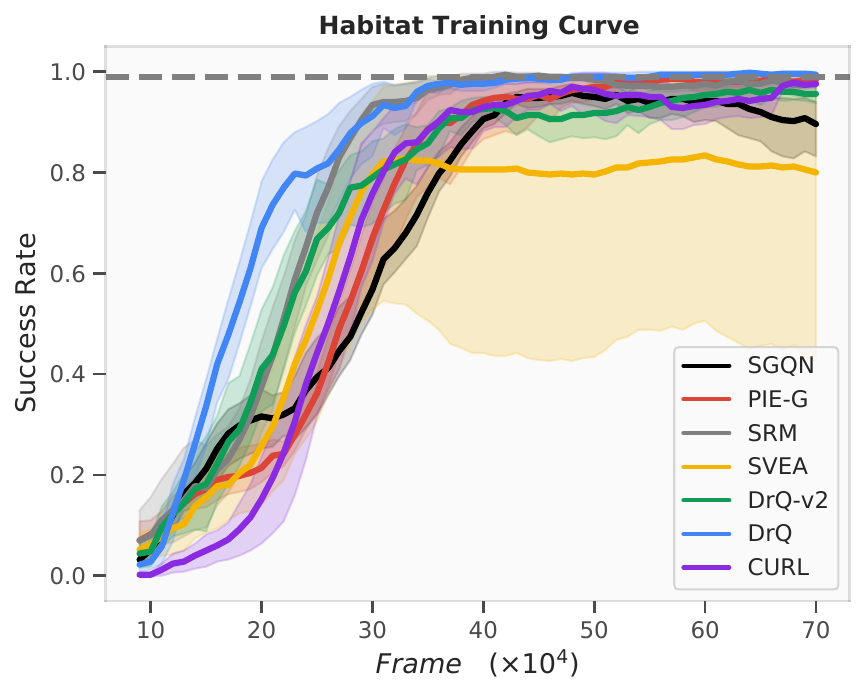}}
    \hspace{6mm}
    \subfloat[ \label{fig:carla_curve}]
        {\includegraphics[width=0.4\linewidth]{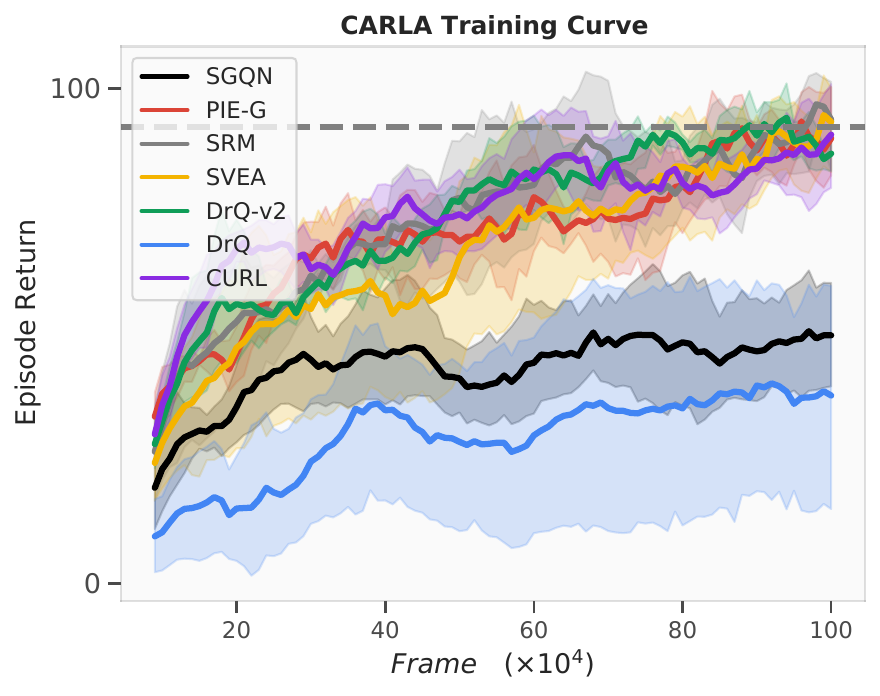}}
\vspace{-5pt}
\caption{\textbf{Sample efficiency of Habitat and CARLA.}  We show the success rate of Habitat and the episode return of CARLA accordingly. The first-person view observations are more robust to the augmentation of adding additional noise.}
\label{fig:habi_carla_sample}
\vspace{-10pt}
\end{figure}

Regarding Habitat and CARLA, as shown in Figure~\ref{fig:habi_carla_sample}, the difference of sample efficiency across diverse algorithms is minimal. This may be attributed to the fact that both two environments employ first-person view rendered images, which makes them more robust to the extra noise. Besides, it should be noted that in CARLA, agents are required to execute fast action changes on the roads to avert collisions with surrounding vehicles. Therefore, Figure~\ref{fig:carla_curve} demonstrates that DrQ is prone to entropy collapse, while SGQN struggles to extract salient information with many distracted factors.

In terms of Adroit, as mentioned in Section~\ref{subsub:dexterous}, the safe Q mechanism is able to endow the trained agent with robustness against noise. The sample efficiency of each algorithm is shown in Figure~\ref{fig:adroit_curve}.

\end{document}